\documentclass{article}

 \usepackage[main, final]{neurips_2026}

\usepackage[utf8]{inputenc} 
\usepackage[T1]{fontenc}    
\usepackage{hyperref}       
\usepackage{url}            
\usepackage{booktabs}       
\usepackage{amsfonts}       
\usepackage{nicefrac}       
\usepackage{microtype}      
\usepackage{xcolor}         

\usepackage{amsmath, amssymb}
\usepackage{amsthm}
\usepackage{graphicx}
\usepackage{xspace}
\usepackage{algpseudocode,algorithm}
\usepackage{multirow}

\newtheorem{theorem}{Theorem}
\newtheorem{remark}{Remark}

\newtheorem{corollary}[theorem]{Corollary}
\newtheorem{proposition}{Proposition}

\newcommand{\modelname}{SGCP\xspace}
\definecolor{mred}{HTML}{b22223}

\title{Stochastic Grouping Conformal Prediction for Effective Subgroup Reliability}

\author{%
  \textbf{Meihui Zhong$^{1}$\thanks{\texttt{mh.zhong1225@outlook.com}}, Wenxin Tai$^{1}$,
  Ting Zhong$^{1,3}$, Fan Zhou$^{1,2,3}$\thanks{Corresponding author}}\\[0.5em]
  $^{1}$ University of Electronic Science and Technology of China\\
  $^{2}$ Intelligent Digital Media Technology, Key Laboratory of Sichuan Province\\
  $^{3}$ Kashi Institute of Electronics and Information Industry
}

\begin{document}

\maketitle

\begin{abstract}
Conformal prediction offers a distribution-free coverage guarantee, making it especially attractive for clinical applications. 
Standard conformal prediction, however, provides such guarantees only at the population level, and its prediction sets can exhibit coverage disparities across clinically important subgroups.
A natural remedy is to calibrate within predefined groups. However, this can require access to sensitive subgroup attributes and is prone to a worst-group bottleneck: protecting the most difficult subgroup can inflate prediction sets for all, increasing cognitive burden on decision makers.
To this end, we propose Stochastic Grouping Conformal Prediction (SGCP), a conformal framework for subgroup-reliable uncertainty quantification. 
It learns a stochastic grouping map that allows each sample to draw calibration information from others with similar calibration behavior, yielding a local score law that boosts reliability across subpopulations.
We prove that SGCP retains the standard coverage guarantee.
Experiments on synthetic and real-world benchmarks show that it consistently reduces subgroup coverage gaps while achieving smaller or comparable prediction set sizes relative to existing baselines.
\end{abstract}

\section{Introduction}
Conformal prediction (CP) has emerged as an attractive framework for medical AI, because it converts any predictive model into a prediction set with finite-sample coverage guarantees under minimal assumptions~\cite{DBLP:conf/aaai/LuLCHK22,cheung2026compass,10.1007/978-3-031-16452-1_52}. 
Consider training data $\{(X_i,Y_i)\}_{i=1}^n$ and a future test point $(X_{n+1},Y_{n+1})$, where each feature $X_i\in\mathcal X$ lies in the feature space and response $Y_i\in\mathcal Y$ lies in the label space.
The goal is to construct a prediction set $\widehat C_n(X_{n+1}) \subseteq \mathcal{Y}$ that contains the unknown test response $Y_{n+1}$ with high probability. 
The appeal of distribution-free prediction is that this guarantee can be obtained without imposing strong assumptions on the underlying data distribution~\cite{Vovk2005, DBLP:journals/jmlr/ShaferV08,10.1561/2200000101}, namely,
\begin{equation}
\mathbb{P}\{Y_{n+1}\in \widehat C_n(X_{n+1})\}\ge 1-\alpha.
\label{eq:intro_marginal}
\end{equation}

While this guarantee is highly appealing, it has an important limitation. 
The coverage in Eq.~\eqref{eq:intro_marginal} is averaged over the calibration data and the future test point, thus providing validity only at the population level~\cite{10.1093/imaiai/iaaa017,NEURIPS2020_244edd7e}.
In clinical settings, this means that a nominal 90\% coverage guarantee may hold on average over all patients, while still failing to provide equally reliable uncertainty estimates for particular demographic groups, disease subtypes, or other clinically meaningful subpopulations. 

This concern is tied to fairness in healthcare, where the costs of unreliable uncertainty often fall disproportionately on underrepresented patient populations~\cite{doi:10.1056/NEJMc2104626, Chen2023}.
Differences in prevalence, data quality, and feature distributions across populations can induce distinct error patterns and nonconformity score distributions~\cite{Kwon2026}. 
Thus, population-level calibration is prone to masking subgroup failures by averaging under-coverage in some groups with over-coverage in others.
In medical applications, this imbalance matters, as what we care about is not only whether uncertainty is valid on average, but whether it remains reliable for the subgroups for whom prediction errors are most costly.

\begin{figure*}[t]
    \centering
    \includegraphics[width=0.957\linewidth]{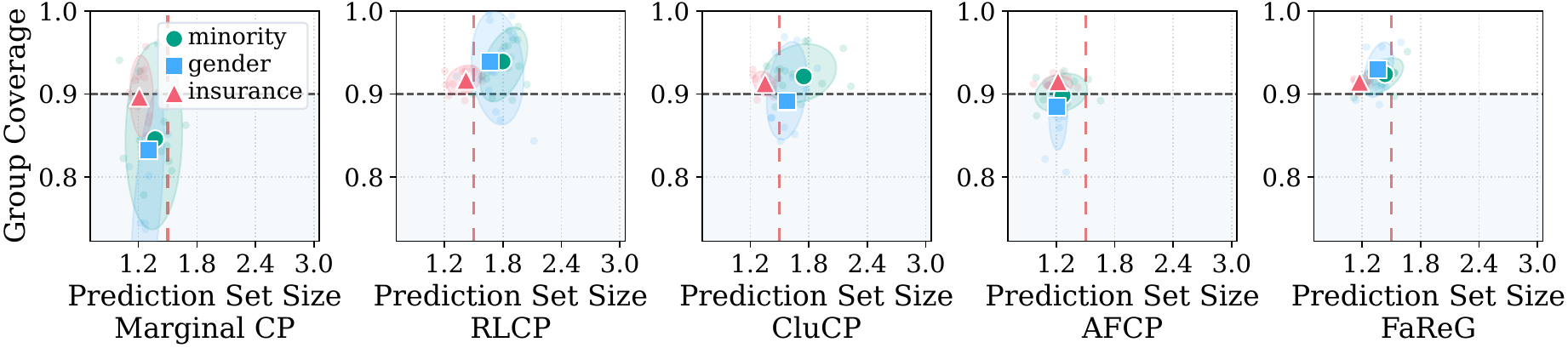}
    \caption{Tradeoff between subgroup coverage and prediction set size on MIMIC-IV~\cite{PhysioNet-mimiciv-3.1} ($1-\alpha=0.9$). Results are shown for RLCP~\citep{10.1093/jrsssb/qkae103}, CluCP~\citep{10.5555/3666122.3668939}, AFCP~\citep{10.5555/3737916.3741370}, and FaReG~\citep{xu2026fair}. The red dashed line marks half of the total number of classes.
    Several fairness-aware CP methods improve sensitive-group coverage, but often do so with much larger prediction sets, making them less informative in practice.}
    \vspace{-1.7mm}
    \label{fig_intro}
\end{figure*}

We empirically observe exactly this tension, as shown in Figure~\ref{fig_intro}. 
Standard conformal prediction achieves the desired overall coverage, but still exhibits lower coverage on several clinically important subpopulations, suggesting that marginal validity may `average out' group-specific uncertainty. 
A natural response is to calibrate separately within each group. 
However, methods~\cite{Romano2020With,xu2026fair} that target finer grained group reliability often become increasingly conservative, resulting in larger prediction sets. 
Such enlarged sets may be less informative in practice and may even hinder downstream decision-making by imposing excessive cognitive burden~\cite{10.1145/3613904.3642446, cresswell2025conformal,cortes-gomez2025utilitydirected}.
One may go even further and calibrate at the individual-sample level, as in localized conformal prediction~\cite{10.1093/jrsssb/qkae103,Romano2020With}. 
Yet our results show that localization, while sometimes improving coverage on subgroups, often incurs substantially larger prediction sets and, more importantly, greater instability across random seeds and data splits.

In sum, these observations pose two coupled challenges.
First, the heterogeneity that matters for calibration is often latent: it may \emph{not be} directly observable at inference, nor align with pre-specified sensitive groups.
Second, the groups that most need tailored calibration are often those with \emph{limited} sample sizes, making fine grained localized calibration conservative and statistically unstable~\citep{10.5555/3666122.3668939}. 
More broadly, we seek a practical middle ground between global calibration, which can wash out subgroup-specific uncertainty, and calibration strategies that rely too heavily on known groups or local neighborhoods, which can become conservative or unstable when calibration data are limited.

Motivated by this gap, we propose \textbf{stochastic grouping conformal prediction} (\modelname), a fairness-aware conformal prediction method that improves subgroup reliability without relying on known sensitive attributes. 
Rather than asking coverage to hold for each individual test point, we relax the requirement to a weaker but still meaningful sense -- coverage over samples that exhibit similar calibration behavior. 
In other words, exact 90\% coverage may be unattainable for one patient in isolation, yet it may still be possible to achieve more reliable coverage on patients whose scores reflect similar uncertainty patterns and calibration difficulty.

To achieve this, \modelname learns a latent calibration structure that organizes samples according to uncertainty-relevant score behavior.
Atop this structure, it learns a \emph{stochastic grouping map} that determines how each sample should draw calibration information from a small set of latent calibration components. 
This yields a sample-adaptive local score law: calibration is tailored to each sample, but estimated through structured sharing rather than from a small group or neighborhood alone.
Statistically, this design strikes a balance between flexibility and stability. 
It is flexible enough to capture heterogeneous score behavior across samples, yet structured enough to be reliably estimated from finite data while preserving conformal validity. 
As a result, \modelname can improve subgroup reliability beyond global calibration without relying on sparse subgroup calibration data. 
Empirically, across fairness-sensitive clinical and public prediction tasks, \modelname improves coverage on clinically meaningful subgroups while maintaining valid overall coverage and efficient prediction sets.

Our contributions are summarized as follows. 
First, we propose \modelname, a novel fairness-aware conformal prediction method. 
It learns a local score law through stochastic grouping, thereby approximating the score distribution of test points. This improves subgroup reliability without relying on rigid predefined groups.
Second, we prove that \modelname retains finite-sample marginal validity. 
We further show that subgroup coverage gaps decrease when the learned local score law better captures the score behavior of meaningful subpopulations.
Third, through experiments on synthetic and medical datasets, we demonstrate that \modelname achieves a favorable tradeoff among overall coverage, subgroup reliability, and prediction set efficiency. We expect \modelname to enable more effective and trustworthy decision-making in high-risk medical applications. 
Code and supplementary materials are included in the anonymous supplementary upload.

\section{Preliminaries} \label{subsec_setup}

We consider multiclass conformal classification with an exchangeable dataset 
$\mathcal D=\{(X_i,Y_i)\}_{i=1}^N$, where $X_i\in\mathcal X$ and $Y_i\in\mathcal Y=\{1,\dots,C\}$. 
Let $f_\theta:\mathcal X\to\Delta^{C-1}$ be a base classifier, with $f_\theta^y(x)$ denoting the predicted probability of class $y$. 
We use the nonconformity score $s(x,y)=1-f_\theta^{\,y}(x)$.
For an observed example $(X_i,Y_i)$, we write $s_i=s(x_i,y_i)$, and it measures how inconsistent a point is with the true label.
Given a training set $\mathcal{D}_{\mathrm{tra}}$ and a calibration set $\mathcal{D}_{\mathrm{cal}}$, we are interested in constructing, for a test point $X_{n+1}$, a confidence set $\widehat{C}(x) \subseteq \mathcal{Y}$ such that $\mathbb P\{Y_{n+1}\in \widehat C(X_{n+1})\}\ge 1-\alpha$, for a miscoverage level $\alpha$, without distributional assumptions beyond exchangeability.

In fairness-sensitive applications, the central issue is not merely whether the prediction set is valid on average, but rather how coverage and set size are distributed across heterogeneous subpopulations.
Ideally, one desire distribution-free \emph{conditional} coverage -- that is, 
\begin{equation}
    \mathbb{P}\{Y \in \widehat{C}(x) \mid X=x\} \ge 1-\alpha \qquad P_X\text{-a.e. } x .
\end{equation}
which would mitigate uneven coverage across patient demographics, disease patterns, and other meaningful groups. 
However, non-trivial distribution-free conditional coverage is impossible, since finite samples cannot ensure tight, valid prediction sets for arbitrary conditioning events~\cite{10.1093/imaiai/iaaa017,10.1093/jrsssb/qkae103}.


We typically classify existing approaches into two broad types.
\textbf{(i) Group-wise calibration.}
A natural response to subgroup miscoverage is to calibrate separately across groups~\cite{DBLP:conf/aaai/LuLCHK22,vovk2003mondrian}, a type we term \emph{Group-wise calibration} CP. 
If $G(x)\in\{1,\dots,K\}$ denotes a group label, this yields a group-specific threshold
$\hat q_k = \mathrm{Quantile}\Bigl(\{s_i:\,(x_i,y_i)\in \mathcal D_{\mathrm{cal}},\ G(x_i)=k\},\,1-\alpha\Bigr)$,
and outputs
\begin{equation}
\widehat C_{\mathrm{GW}}(x) =
\{y\in\mathcal Y:\ s(x,y)\le \hat q_{G(x)}\}.
\label{eq_groupwise}
\end{equation}
In principle, Eq.~\eqref{eq_groupwise} reduces the mismatch induced by a single global threshold.
In practice, however, one issue is that the relevant groups are often not known a priori, and may not align with pre-specified sensitive attributes.
Another is that finer grouping fragments the calibration set, leaving some groups with too few samples for calibration, and making the empirical quantiles $\hat q_k$ statistically unstable~\citep{10.5555/3666122.3668939}.

This group-wise perspective also includes methods that learn or adapt groups before calibration. 
AFCP~\citep{10.5555/3737916.3741370}, for example, considers groups defined by sensitive attributes and makes the group selection itself adaptive.
Let $\phi(x,\mathcal A)$ denote the projection of $x$ onto a selected sensitive-attribute set $\mathcal A$. For instance, a subgroup ``males with a bachelor's degree'' can be written as $\phi(x,\{k,\ell\})\in[M_k]\times[M_\ell]$. 
For each test point, AFCP selects either one sensitive attribute or none, $\hat A(X_{n+1})\in\{\emptyset,\{1\},\dots,\{K\}\}$, and targets $\mathbb P\!\left[Y_{n+1}\in \hat C(X_{n+1})
\mid \phi \bigl(X_{n+1},\hat A(X_{n+1})\bigr)
\right] \ge 1-\alpha$.
This reduces to marginal coverage when $\hat A(X_{n+1})=\emptyset$.
Operationally, AFCP combines a marginal conformal set with an equalized-coverage set for each candidate label under the selected attribute. 
Although less conservative than exhaustive calibration over all attribute combinations, AFCP remains tied to discrete sensitive attributes. 
Its repeated leave-one-out calibration on pseudo-test points can also be computationally expensive and noisy in sparse-group regimes.

FAREG~\citep{xu2026fair} further improves group discovery by learning groups in a latent representation space rather than enumerating them from raw input features. This expands the class of detectable subgroups, including those defined by nonlinear feature combinations. However, FAREG still frames fairness as subgroup discovery: it identifies explicit subgroups with coverage disparity and enforces coverage over them. 
It is therefore most suitable when heterogeneity is well captured by identifiable subgroups. It may, however, become conservative in some cases.


\textbf{(ii) Localized calibration.}
Localized calibration, as in RLCP~\citep{10.1093/jrsssb/qkae103}, avoids explicit group partitions by using nearby samples as the reference set for each test point. 
RLCP argues that heterogeneity relevant to calibration may be highly sample-specific.
Let $\mathfrak X_x=\{\mathcal X\in\mathfrak X:\,x\in\mathcal X\}$ denote the candidate neighborhoods containing $x$.
A localized threshold can be written as
$\hat q_{\mathrm{loc}}(x) = \sup_{\mathcal X\in \mathfrak X_x}
\mathrm{Quantile}\Bigl(
\{s_i:\,(x_i,y_i)\in\mathcal D_{\mathrm{cal}},\ x_i\in\mathcal X\},
\,1-\alpha
\Bigr)$,
with prediction set
\begin{equation}
\widehat C_{\mathrm{loc}}(x) =
\{y\in\mathcal Y:\ s(x,y)\le \hat q_{\mathrm{loc}}(x)\}.
\label{eq_localized}
\end{equation}
Compared with group-wise calibration, Eq.~\eqref{eq_localized} avoids rigid group boundaries and adapts to the region around the test point. 
Yet the supremum over candidate neighborhoods requires the prediction set to cover the worst local subset containing $x$, which may be overly conservative and yield unnecessarily large sets, particularly when calibration samples are scarce. 
Moreover, locality is typically defined via feature space similarity, which need not align with the actual error structure most relevant for calibration.
RLCP may introduces additional randomness into the prediction intervals, leading to inconsistent outputs across independent runs.
Other related works are discussed in Appendix~\ref{appendix_rw}.

The preceding discussion suggests that the key challenge is to improve subgroup reliability without losing statistical stability or producing overly large prediction sets. We therefore seek a method that can capture calibration heterogeneity while promoting informative prediction sets.

\begin{figure*}[t]
    \centering
    \includegraphics[width=0.85\linewidth]{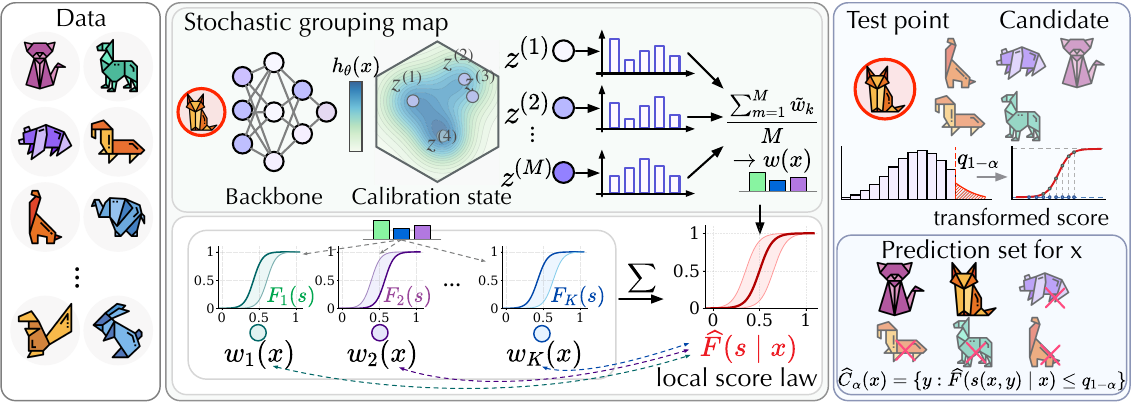}
    \caption{Overview of SGCP. SGCP learns a local score law by mapping $x$ to a stochastic grouping map $w(x)$ over latent calibration components $\{F_k\}_{k=1}^K $. Candidate labels are transformed by this local law and kept when their calibrated scores fall below the conformal threshold.}
    \label{fig_method}
\end{figure*}

\section{Method} \label{sec_method}

\subsection{From subgroup reliability to local score laws}
Our starting point is that conformal calibration is ultimately governed by the distribution of the nonconformity score. 
Accordingly, we take the sample-specific score distribution $F_x^\star(s):=\mathbb P(S\le s\mid X=x)$, where $S$ denotes the nonconformity score, as the oracle object for reliable calibration to each subgroup.
If we could estimate $F_x^\star$ accurately, then each sample could be calibrated according to its own local score law.
That said, the difficulty is that directly learning an unconstrained conditional distribution $F_x^\star$ for every $x$ is data-hungry and unstable in finite samples, especially when the relevant heterogeneity is latent and high-dimensional.
A single test point rarely provides enough local calibration evidence for reliable estimation.
This motivates the central question behind \modelname: can other samples help a target point estimate its local score distribution without resorting to a single global calibration?
\modelname addresses this by replacing the unconstrained family $\{F_x^\star:x\in\mathcal X\}$ with a structured mixture family.
Concretely, \modelname approximates the oracle local score law by
\begin{equation}
    F_x^{\star}(s)\approx 
    \widehat F(s\mid x)=\sum_{k=1}^K w_k(x)\,F_k(s),
    \label{eq_oracle_score_law}
\end{equation}
where $F_1,\dots,F_K$ are \emph{latent calibration components} and $w(x)$ is a sample-related \emph{stochastic grouping map}.
Thus, each sample has its own effective local score law, but that law is not estimated from scratch. It is represented as a weighted combination of latent calibration components learned from related samples.
This makes \modelname more expressive than discrete group-wise calibration, while still allowing samples with similar calibration behavior to share evidence through the components. Posterior averaging further stabilizes the resulting local score law (see Appendix~\ref{appendx_consistency} for details). 
Figure~\ref{fig_method} illustrates how stochastic grouping induces a local score law for each sample.

The key question in Eq.~\eqref{eq_oracle_score_law} is therefore how to determine, for each sample $x$, which other samples should be regarded as `related'.
While there are many ways to define relevance around a point, such as finding similar points through feature embedding or clustering, recall that the ideal object for subgroup reliable calibration is the sample-specific score distribution $F_x^\star$ in Eq.~\eqref{eq_oracle_score_law}.
Accordingly, the relevant notion of locality is not simply geometric proximity in the input space, but similarity in calibration relevant score behavior.
So we want to identify samples that are governed by similar score behavior.
This motivates us to propose the latent calibration state and stochastic grouping map. Together, they determine how each sample should transfer calibration evidence from the latent calibration components.
We next formalize these two components and the resulting local score law.

\subsection{Learning a differentiable local score law}

\paragraph{Latent calibration state.}
Let $h_\theta(x)\in\mathbb R^d$ denote the feature representation produced by the backbone classifier, and let $f_\theta(x)\in\Delta^{C-1}$ denote the corresponding predictive probability vector.
To capture heterogeneity in score behavior beyond predefined groups, we associate each sample with a latent calibration state $z$.
We model this state via a posterior distribution
\begin{equation}
q_\phi(z\mid x)
=
\mathcal N\!\Bigl(
\mu_\phi\bigl(h_\theta(x)\bigr),
\operatorname{diag}\bigl(\sigma_\phi^2(h_\theta(x))\bigr)
\Bigr),
\label{eq:posterior}
\end{equation}
where $\mu_\phi(\cdot)$ and $\sigma_\phi(\cdot)$ are learned functions of the backbone representation.
This stochastic formulation allows calibration behavior to remain ambiguous even when the observed feature vector is fixed.
Samples with similar predictions may still display mixed or uncertain score behavior, and such ambiguity is naturally represented through posterior uncertainty in latent space.

\paragraph{Stochastic grouping map.}
For a latent draw $z\sim q_\phi(z\mid x)$, we obtain a simplex-valued membership vector $\tilde w(x,z)=\bigl(\tilde w_1(x,z),\dots,\tilde w_K(x,z)\bigr)\in\Delta^{K-1}$ through a learnable differentiable mapping from the latent state to the probability simplex.
A single latent draw represents only one possible realization of the posterior calibration state and therefore only one possible decomposition of the sample's local score law.
What we seek, however, is a stable description of how the sample is associated with the latent calibration components.
Accordingly, \modelname defines the \emph{stochastic grouping map} by averaging memberships over the posterior distribution.
Let $z^{(1)},\dots,z^{(M)} \overset{\mathrm{i.i.d.}}{\sim} q_\phi(z\mid x)$.
We define the posterior-averaged membership of sample $x$ on the $k$-th latent calibration component
\begin{equation}
w_k(x)
=
\mathbb E_{z\sim q_\phi(\cdot\mid x)}[\tilde w_k(x,z)]
\approx
\frac{1}{M}\sum_{m=1}^M \tilde w_k\bigl(x,z^{(m)}\bigr),
\qquad k=1,\dots,K.
\label{eq_avg_weights}
\end{equation}
This yields the stochastic grouping map
$w(x)=\bigl(w_1(x),\dots,w_K(x)\bigr)\in\Delta^{K-1}$, which determines how the latent calibration components are combined to form the learned local score law.

\paragraph{Aggregate local score law.}
Each latent calibration component $F_k:[0,1]\to[0,1]$ is a monotone mapping that represents one recurring calibration pattern in the score space.
When needed, we write $f_k$ for the corresponding density.
Given the membership weight $w_k(x)$, \modelname defines the aggregate local score law at $x$ as
\begin{equation}
\widehat F(s\mid x)
=
\sum_{k=1}^K w_k(x)\,F_k(s),
\label{eq_local_score_law}
\end{equation}
with corresponding local density
$\widehat f(s\mid x) = \sum_{k=1}^K w_k(x)\,f_k(s)$.
Equation~\eqref{eq_local_score_law} defines the central local score-law estimator in \modelname.
It replaces the infeasible task of estimating an unconstrained $F_x^\star$ for every test point with a structured family generated by a small number of latent calibration components.
In our implementation, each $F_k$ is parameterized as a monotone differentiable mapping.

We next explain why local score-law normalization matters for subgroup reliability. 
If the local score law were known, the probability integral transform would map true-label nonconformity score onto a common percentile scale.
The following proposition formalizes this property (proof in Appendix~\ref{appendix_proof_scp}).

\begin{proposition}\label{prop_oracle}
Let $V(x,y):=\widehat F\bigl(s(x,y)\mid x\bigr)$ denote the transformed score.
Suppose that the learned local score law coincides with the oracle local score law, namely, for $P_X$-almost every $x$, $\widehat F(t\mid x)=F_x^\star(t):=\mathbb P(S\le t\mid X=x)$, $\forall t\in[0,1]$.
Assume furthermore that, for $P_X$-a.e. $x$, the conditional score distribution $S\mid X=x$ is continuous. Then, for any measurable subgroup $G\subseteq\mathcal X$ with $\mathbb P(X\in G)>0$,
\[
V(X,Y)\mid \{X\in G\}\sim \mathrm{Unif}(0,1).
\]
\end{proposition}


Proposition~\ref{prop_oracle} explains why local score law learning can remove subgroup reliability inconsistency in the oracle case.
In practice, \modelname learns this transform approximately, using the following objective.

\paragraph{Learning objective.}
We split the training data into $\mathcal D_{\mathrm{tra}}=\mathcal D_0\cup\mathcal D_1$.
The split $\mathcal D_0$ is used to learn the backbone predictor, the latent-state posterior, and the stochastic grouping map, while $\mathcal D_1$ is used to fit the latent calibration components and the learned local score law. An additional independent calibration set $\mathcal D_{\mathrm{cal}}$ is reserved for split conformal calibration.
For each $(X_i,Y_i)\in\mathcal D_1$, we compute the true-label nonconformity score $s_i=s(X_i,Y_i)$, the membership weights $w_{ik}=w_k(X_i)$, and the component-wise transformed score $C_{ik}=F_k(s_i)$. Since $F_k$ is monotone, $C_{ik}$ can be viewed as the percentile position of $s_i$ under the $k$-th latent calibration component.

The training objective encourages three complementary properties.
First, within each component, the weighted distribution of transformed scores should be close to $\mathrm{Unif}(0,1)$. This yields the rank-uniformization loss $\mathcal L_{\mathrm{rank}}$.
Second, the induced local mixture density $\widehat f(s_i\mid X_i)=\sum_{k=1}^K w_{ik} f_k(s_i)$ should explain the observed true-label scores, yielding the density-fitting loss $\mathcal L_{\mathrm{nll}}$.
Finally, the stochastic grouping map should be informative: it should use the latent components diversely across the population while retaining sample-specific membership weights.
We encourage this by maximizing the empirical mutual information between samples and a latent component variable $Z$, denoted by $\widehat I^{\,\mathrm{norm}}(X;Z)$, where the membership weights define $q_\phi(Z=k\mid X_i)=w_{ik}$.

Combining these terms, we optimize
\begin{equation}
\mathcal L_{\mathrm{\modelname}}
=
\mathcal L_{\mathrm{rank}}
+
\lambda\mathcal L_{\mathrm{nll}}
-
\beta \widehat I^{\,\mathrm{norm}}(X;Z),
\label{eq_loss}
\end{equation}
where $\lambda$ controls the strength of local density fitting and $\beta$ controls the strength of stochastic grouping information maximization.
Detailed parameterizations and explicit loss definitions are deferred to Appendix~\ref{app_implementation_sgcp}.
Algorithm~\ref{alg:sgcp} summarizes the overall procedure.

\begin{algorithm}[t]
\caption{Procedure for SGCP}
\label{alg:sgcp}
\small
\begin{algorithmic}[1]
\Require Training data $\mathcal D_{\mathrm{tra}}$, calibration data $\mathcal D_{\mathrm{cal}}$, test input $X_{n+1}$, nonconformity score $s(\cdot,\cdot)$, $K$ latent calibration components, $M$ posterior samples, hyperparameters $\lambda, \beta$, miscoverage level $\alpha$
\Ensure Prediction set $\widehat C_\alpha(X_{n+1})$
\State Split $\mathcal D_{\mathrm{tra}}$ into $\mathcal D_0$ and $\mathcal D_1$
\State \textbf{Fit \modelname:} on $\mathcal D_0$, learn $f_\theta$, latent-state posterior $q_\phi(z\mid x)$ in Eq.~\eqref{eq:posterior}, and stochastic grouping map $w(x)$ in Eq.~\eqref{eq_avg_weights}; on $\mathcal D_1$, fit latent calibration components $\{F_k\}_{k=1}^K$ and local score law $\widehat F(\cdot\mid x)$ in Eq.~\eqref{eq_local_score_law}.
\State \textbf{Calibration:} for each $(X_i,Y_i)\in\mathcal D_\mathrm{cal}$, compute $s_i=s(X_i,Y_i)$ and $V_i=\widehat F(s_i\mid X_i)$.

\State \textbf{Quantile:} sort $\{V_i:(X_i,Y_i)\in\mathcal D_\mathrm{cal}\}$ as $V_{(1)}\le\cdots\le V_{(m)}$, where $m=|\mathcal D_\mathrm{cal}|$, and set
$\widehat q_{1-\alpha}=V_{\left(\lceil (m+1)(1-\alpha)\rceil\right)}$, with fallback $\widehat q_{1-\alpha}=1$ if $\lceil (m+1)(1-\alpha)\rceil>m$.

\State \textbf{Test transform:} for each candidate label $y\in\mathcal Y$, compute
$V_{n+1}^{\,y}
=
\widehat F\bigl(s(X_{n+1},y)\mid X_{n+1}\bigr)$.

\State \textbf{Return:}
$\widehat C_\alpha(X_{n+1})
=
\left\{
y\in\mathcal Y:\;
V_{n+1}^{\,y}\le \widehat q_{1-\alpha}
\right\}$.
\end{algorithmic}
\end{algorithm}

The following result, proved in Appendix~\ref{proof_validity}, shows that \modelname retains the standard finite-sample marginal validity guarantee of split conformal prediction, which contains the true label with probability at least $1-\alpha$.
This guarantee does not depend on how accurately the local score law is learned, and the prediction set $\widehat C_\alpha(X_{n+1})$ remains valid.

\begin{proposition}\label{prop_validity}
Under the standard split-conformal exchangeability assumption, the prediction set returned by \modelname\ satisfies $\mathbb P\!\left\{
Y_{n+1}\in \widehat C_\alpha(X_{n+1})
\right\}
\ge 1-\alpha$.
\end{proposition}

Marginal validity alone does not explain subgroup reliability.
A further question is how the learned transform affects subgroup coverage.
The following (proved in Appendix~\ref{proof_approx_group_coverage}) shows that subgroup coverage is controlled by the mismatch between subgroup and overall transformed-score distributions.
In the oracle property of Proposition~\ref{prop_oracle}, this mismatch vanishes for every subgroup. In practice, \modelname seeks to reduce it by learning a local score law transform through stochastic grouping. Appendix~\ref{cor_lowerbound} shows that uniformly approximating the oracle score law directly controls this mismatch.


\begin{proposition}
\label{prop_approx_group_coverage}
For any measurable subgroup $G$ with $\mathbb P(X\in G)>0$, let $\Delta_G$ denote the Kolmogorov distance between the distribution of $V(X,Y)$ conditional on $X\in G$ and its marginal distribution.
Under the assumptions of Proposition~\ref{prop_validity} and a continuous marginal distribution of $V(X,Y)$, subgroup coverage is lower bounded by
$\mathbb P\!\left\{
Y_{n+1}\in \widehat C_\alpha(X_{n+1})
\mid
X_{n+1}\in G
\right\}
\ge
1-\alpha-\Delta_G
$.
\end{proposition}

\section{Numerical Experiments} \label{sec_exp}

\paragraph{Benchmarks and preprocessing.}
We evaluate \modelname on four benchmarks.
For each dataset, let $(X,Y,A)$ denote the feature vector, class label, and sensitive attributes, respectively. Below, we briefly describe how each dataset is processed and constructed for our evaluation.

\textbf{(i)}
We generate synthetic (SYN) multiclass cardiopulmonary diagnosis benchmark with six disease categories, viz. Hypertension, Coronary Artery Disease, Heart Failure, Arrhythmia, Valvular Heart Disease, and Peripheral Artery Disease. 
Each sample contains four grouping attributes--gender, risk phenotype, age, and region--and six non-sensitive physiological variables. 
To mimic intersectional heterogeneity, individuals satisfying the XNOR relation between biological sex and risk phenotype, e.g., Female with Type I or Male with Type II, are assigned higher label uncertainty. 
This subgroup is therefore harder to classify and serves as the disadvantaged population.
\textbf{(ii)}
Nursery~\cite{nursery_76} is an admission-decision dataset for ranking nursery school applications, where family, socioeconomic, social, and health-related factors make it well suited for evaluating subgroup reliability.
We use four priority ranks as targets and five categorical grouping attributes.
Following~\citep{10.5555/3737916.3741370,xu2026fair}, we simulate controllable algorithmic bias by downsampling the subgroup defined by \texttt{parents} to 10\%, adding label noise, and rounding labels to the nearest valid class.
\textbf{(iii)}
MIMIC-IV~\cite{PhysioNet-mimiciv-3.1} is a real-world clinical tabular benchmark.
We employ three binary grouping attributes, i.e., race, gender, and insurance status, for subgroup analysis, along with non-sensitive demographic and administrative covariates. 
We study two tasks: in-hospital mortality prediction and ICU-delay classification, where delays are grouped into short ($\le4$ hours), moderate ($4$-$12$ hours), and prolonged ($>12$ hours) or no ICU admission.
\textbf{(iv)}
BACH~\cite{ARESTA2019122} is a four-class histopathology benchmark for breast cancer diagnosis, viz. Normal, Benign, In Situ Carcinoma, and Invasive Car2cinoma. 
Since demographic attributes are unavailable, we use the class label as a proxy grouping variable, setting $A:=Y$.
Images are encoded by a visual backbone as $X_i=\Phi(I_i)\in\mathbb R^d$.
Subgroup heterogeneity arises from visual variability across tissue and potential acquisition bias~\cite{9525406}, so the class-conditional score laws $F_c(s):=\mathbb P(S\le s\mid Y=c)$ may differ substantially across classes.

Across all datasets, we randomly split data into training, calibration, and test sets; for \modelname, the training set is further divided equally into $\mathcal D_0$ and $\mathcal D_1$.
Sensitive attributes are used only for fairness evaluation and never as model inputs.
Preprocessing details are given in Appendix~\ref{appendix_experiment}.



We benchmark \modelname against AFCP~\citep{10.5555/3737916.3741370}, Clustered CP~\citep{10.5555/3666122.3668939}, FaReG~\citep{xu2026fair}, RLCP~\citep{10.1093/jrsssb/qkae103}, marginal CP, Partial Equalized~\citep{Romano2020With}, and Exhaustive Equalized. 
Partial Equalized protects one sensitive attribute at a time and unions the resulting sets, while Exhaustive Equalized protects all candidate attributes jointly as an explicit subgroup-equalization baseline.
All methods use the same backbone and nonconformity score. Implementation details and hyperparameters are provided in Appendix~\ref{appendix_para}. See the Appendix~\ref{appendix_exp} for additional experiments.

\paragraph{Evaluation criteria.}
We evaluate all methods in terms of overall validity, subgroup reliability, and efficiency.
Let $\{(X'_i,Y'_i)\}_{i=1}^{N'}$ be the test set, and let $\mathcal C(X'_i)$ denote the prediction set at miscoverage level $\alpha$.
We report empirical \emph{overall coverage}
$\mathrm{Cov.} = \frac{1}{N'}\sum_{i=1}^{N'} \mathbf 1\{Y'_i\in\mathcal C(X'_i)\}$,
which should be close to the target level $1-\alpha$.
Efficiency is measured by the \emph{average set size}
$\mathrm{AvgSize} =
\frac{1}{N'}\sum_{i=1}^{N'} |\mathcal C(X'_i)|$.
Among methods with comparable coverage, smaller AvgSize is preferable.




Given a grouping attribute $A$ and its induced groups, we are interested in coverage conditional on each group. 
For group $g\in\mathcal G(A)$, let $\mathcal I_g:=\{i: A(X'_i)=g\}$. 
We define the estimated conditional coverage, hereafter \emph{group coverage}, as
$\hat c_g =
\frac{1}{|\mathcal I_g|}
\sum_{i\in \mathcal I_g}\mathbf 1\{Y'_i\in\mathcal C(X'_i)\}
$.
We quantify subgroup disparity by the \emph{coverage gap} (CovGap)
$
\mathrm{CovGap}(A) = \frac{1}{|\mathcal G(A)|} \sum_{g\in\mathcal G(A)}
\left|
\hat c_g-(1-\alpha)
\right|
$.
Group coverage closer to $1-\alpha$ and smaller CovGap indicate more reliable subgroup performance.
All metrics other than AvgSize are reported as percentages.

\subsection{Results and Discussion}

\begin{table*}[t]
\centering
\caption{Each method is run 20 times, and we report the mean and standard deviation. All results are reported for the largest sample size. 
Results with nominal coverage are shown in \textcolor{blue}{blue}, optimal values are \textbf{boldfaced}, and AvgSize exceeding half of the total number of categories are highlighted in \textcolor{mred}{red}.}
\label{tab_main_results}
\setlength{\tabcolsep}{2pt} 
\renewcommand{\arraystretch}{1.3}
\resizebox{\linewidth}{!}{
\begin{tabular}{l ccc ccc ccc ccc}
\toprule
\multirow{2}{*}{\textbf{Method}} & \multicolumn{3}{c}{\textbf{SYN}} & \multicolumn{3}{c}{\textbf{Nursery}} & \multicolumn{3}{c}{\textbf{MIMIC-IV}} & \multicolumn{3}{c}{\textbf{BACH}} \\
\cmidrule(lr){2-4} \cmidrule(lr){5-7} \cmidrule(lr){8-10} \cmidrule(lr){11-13}
 & Cov. $\uparrow$ & AvgSize $\downarrow$ & CovGap $\downarrow$ & Cov. $\uparrow$ & AvgSize $\downarrow$ & CovGap $\downarrow$ & Cov. $\uparrow$ & AvgSize $\downarrow$ & CovGap $\downarrow$ & Cov. $\uparrow$ & AvgSize $\downarrow$ & CovGap $\downarrow$ \\
\midrule
Marginal    & 89.52$_{\pm 1.52}$ & 2.309$_{\pm 0.078}$ & 14.5$_{\pm 1.6}$ & 84.15$_{\pm 5.20}$ & \textcolor{mred}{2.170}$_{\pm 0.260}$ & 20.4$_{\pm 5.2}$ & \textcolor{blue}{90.79$_{\pm 10.62}$} & 1.226$_{\pm 0.087}$ & 27.1$_{\pm 7.1}$ & 89.20$_{\pm 9.64}$ & \textcolor{mred}{2.894$_{\pm 0.267}$} & 15.7$_{\pm 1.7}$ \\
Partial     & \textcolor{blue}{94.50$_{\pm 1.12}$} & 2.419$_{\pm 0.048}$ & 6.2$_{\pm 0.7}$ & 88.97$_{\pm 3.80}$ & 1.850$_{\pm 0.200}$ & 12.1$_{\pm 2.5}$ & \textcolor{blue}{94.50$_{\pm 4.54}$} & 1.258$_{\pm 0.140}$ & 8.0$_{\pm 3.0}$ & \textcolor{blue}{96.57$_{\pm 8.08}$} & \textcolor{mred}{3.385$_{\pm 0.299}$} & 7.6$_{\pm 1.6}$ \\
Exhaustive  & \textcolor{blue}{95.85$_{\pm 0.91}$} & \textcolor{mred}{3.342$_{\pm 0.053}$} & 6.0$_{\pm 0.6}$ & \textcolor{blue}{96.80$_{\pm 3.10}$} & \textcolor{mred}{2.850$_{\pm 0.220}$} & 15.5$_{\pm 3.2}$ & \textcolor{blue}{96.45$_{\pm 4.22}$} & \textcolor{mred}{2.800$_{\pm 0.142}$} & 19.7$_{\pm 3.2}$ & \textcolor{blue}{97.41$_{\pm 6.89}$} & \textcolor{mred}{3.867$_{\pm 0.118}$} & 11.6$_{\pm 1.5}$ \\
RLCP        & \textcolor{blue}{93.10$_{\pm 2.45}$} & 2.540$_{\pm 0.305}$ & 4.5$_{\pm 1.8}$ & 88.67$_{\pm 6.20}$ & 1.220$_{\pm 0.410}$ & 13.5$_{\pm 3.8}$ & \textcolor{blue}{92.50$_{\pm 2.80}$} & 1.350$_{\pm 0.100}$ & 1.7$_{\pm 0.8}$ & \textcolor{blue}{91.88$_{\pm 10.56}$} & \textcolor{mred}{2.311$_{\pm 0.319}$} & 8.1$_{\pm 2.5}$ \\
CluCP       & 89.80$_{\pm 1.78}$ & 2.426$_{\pm 0.050}$ & 4.9$_{\pm 0.8}$ & 85.08$_{\pm 1.60}$ & 1.121$_{\pm 0.064}$ & 12.8$_{\pm 1.3}$ & \textcolor{blue}{92.22$_{\pm 1.28}$} & 1.302$_{\pm 0.059}$ & 3.5$_{\pm 0.8}$ & \textcolor{blue}{95.91$_{\pm 3.59}$} & \textcolor{mred}{3.257$_{\pm 0.220}$} & 10.7$_{\pm 2.6}$ \\
AFCP        & \textcolor{blue}{91.77$_{\pm 1.23}$} & 2.432$_{\pm 0.043}$ & 4.6$_{\pm 0.5}$ & \textcolor{blue}{92.85$_{\pm 0.84}$} & 1.085$_{\pm 0.024}$ & 14.1$_{\pm 0.6}$ & - & - & - & \textcolor{blue}{95.00$_{\pm 2.60}$} & \textcolor{mred}{2.830$_{\pm 0.237}$} & 6.3$_{\pm 1.7}$ \\
FaReG       & \textcolor{blue}{92.61$_{\pm 0.98}$} & 2.579$_{\pm 0.029}$ & 4.7$_{\pm 0.5}$ & \textcolor{blue}{93.21$_{\pm 0.62}$} & 1.230$_{\pm 0.012}$ & 13.3$_{\pm 0.6}$ & \textcolor{blue}{90.23$_{\pm 0.09}$} & 1.174$_{\pm 0.001}$ & \textbf{0.5$_{\pm 0.0}$} & \textcolor{blue}{93.60$_{\pm 3.40}$} & \textcolor{mred}{2.959$_{\pm 0.239}$} & 7.4$_{\pm 1.5}$ \\
\midrule
\textbf{\modelname} & \textbf{\textcolor{blue}{90.13$_{\pm 0.39}$}} & \textbf{2.225$_{\pm 0.035}$} & \textbf{4.2$_{\pm 0.7}$} & \textbf{\textcolor{blue}{91.65$_{\pm 0.70}$}} & 1.089$_{\pm 0.035}$ & \textbf{11.8$_{\pm 0.7}$} & \textbf{\textcolor{blue}{90.13$_{\pm 0.07}$}} & \textbf{1.107$_{\pm 0.001}$} & 0.8$_{\pm 0.0}$ & \textbf{\textcolor{blue}{91.90$_{\pm 2.70}$}} & \textbf{\textcolor{mred}{2.224$_{\pm 0.215}$}} & \textbf{6.2$_{\pm 1.1}$} \\
\bottomrule
\end{tabular}
}
\end{table*}

\begin{figure}[t]
  \centering
  \begin{minipage}[t]{0.24\linewidth}
    \centering
    \includegraphics[width=\linewidth]{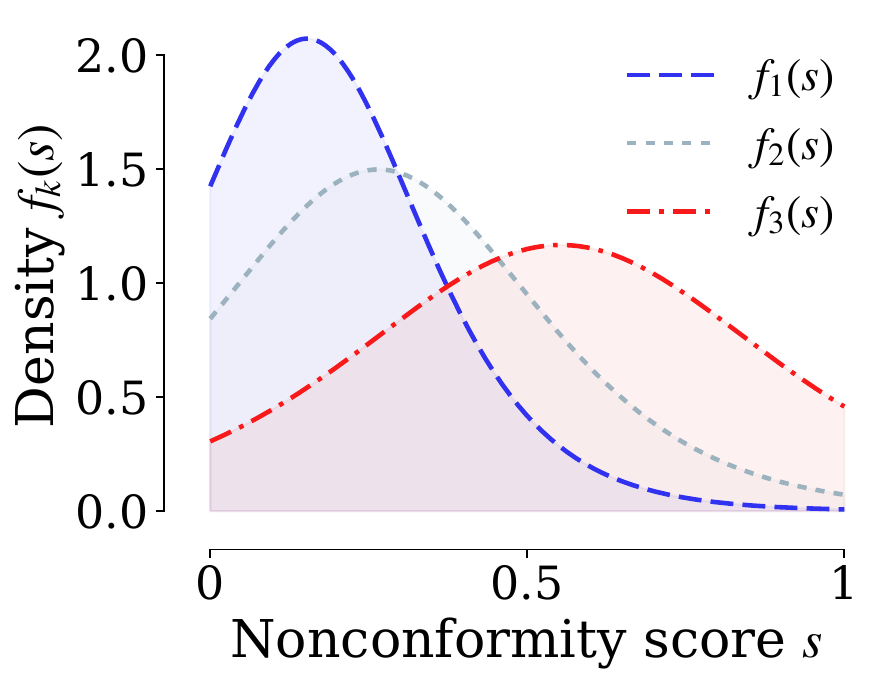}
    {\small (a) SYN}
  \end{minipage}
  \hfill
  \begin{minipage}[t]{0.23\linewidth}
    \centering
    \includegraphics[width=\linewidth]{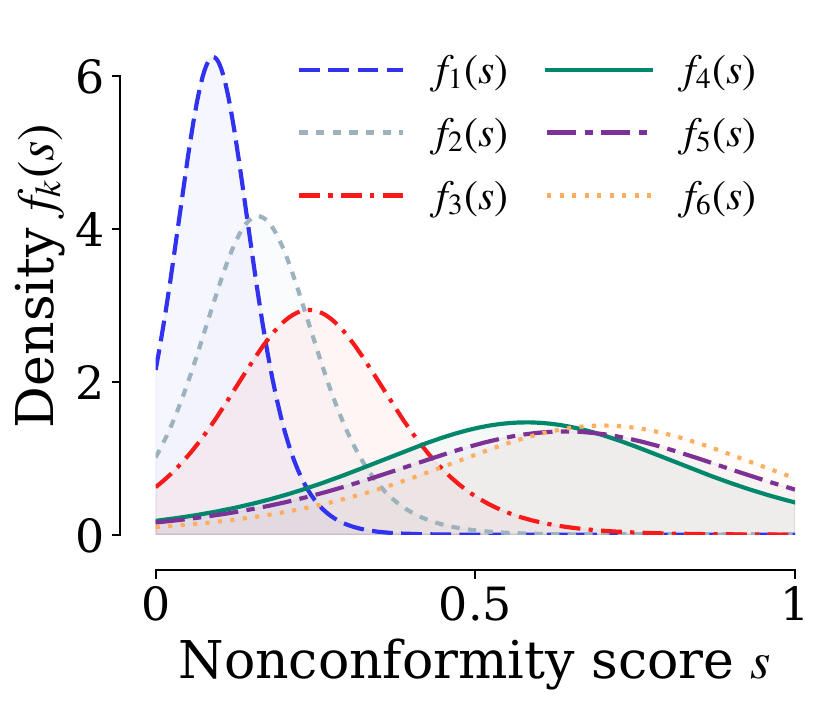}
    {\small (b) Nursery}
  \end{minipage}
  \hfill
  \begin{minipage}[t]{0.24\linewidth}
    \centering
    \includegraphics[width=\linewidth]{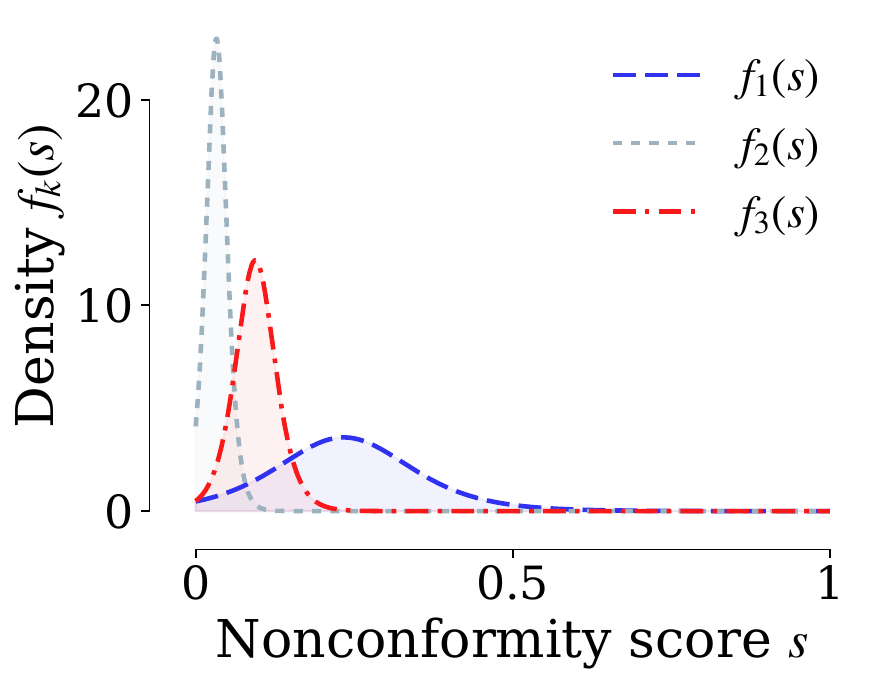}
    {\small (c) MIMIC-IV}
  \end{minipage}
  \hfill
  \begin{minipage}[t]{0.24\linewidth}
    \centering
    \includegraphics[width=\linewidth]{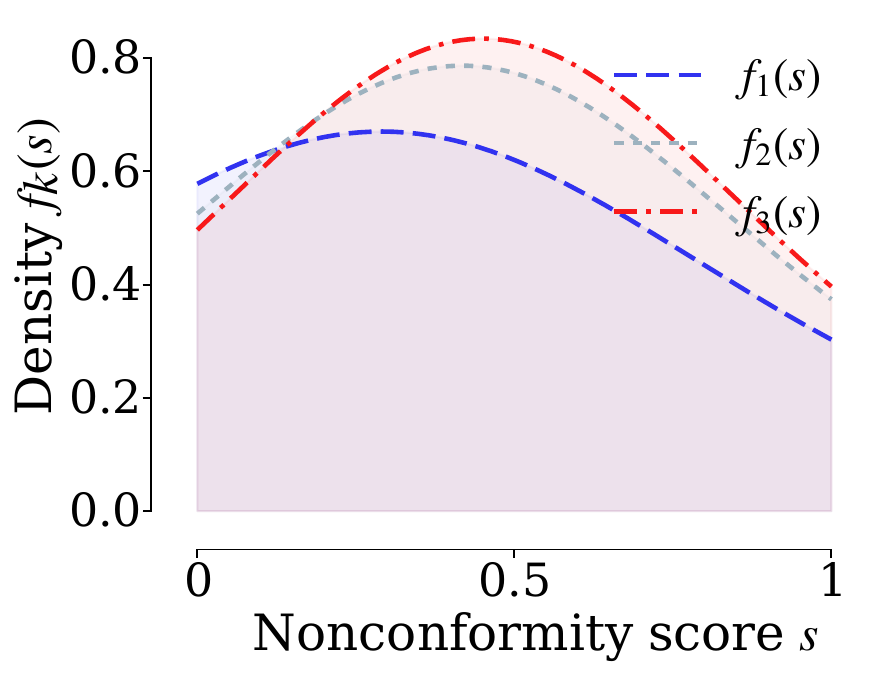}
    {\small (d) BACH}
  \end{minipage}
    \caption{Learned latent component densities on four datasets. The components reveal distinct calibration regimes in the nonconformity-score space.}
    \vspace{-2mm}
  \label{fig_pdf}
\end{figure}

\begin{figure}[t]
    \centering
    \includegraphics[width=0.82\linewidth]{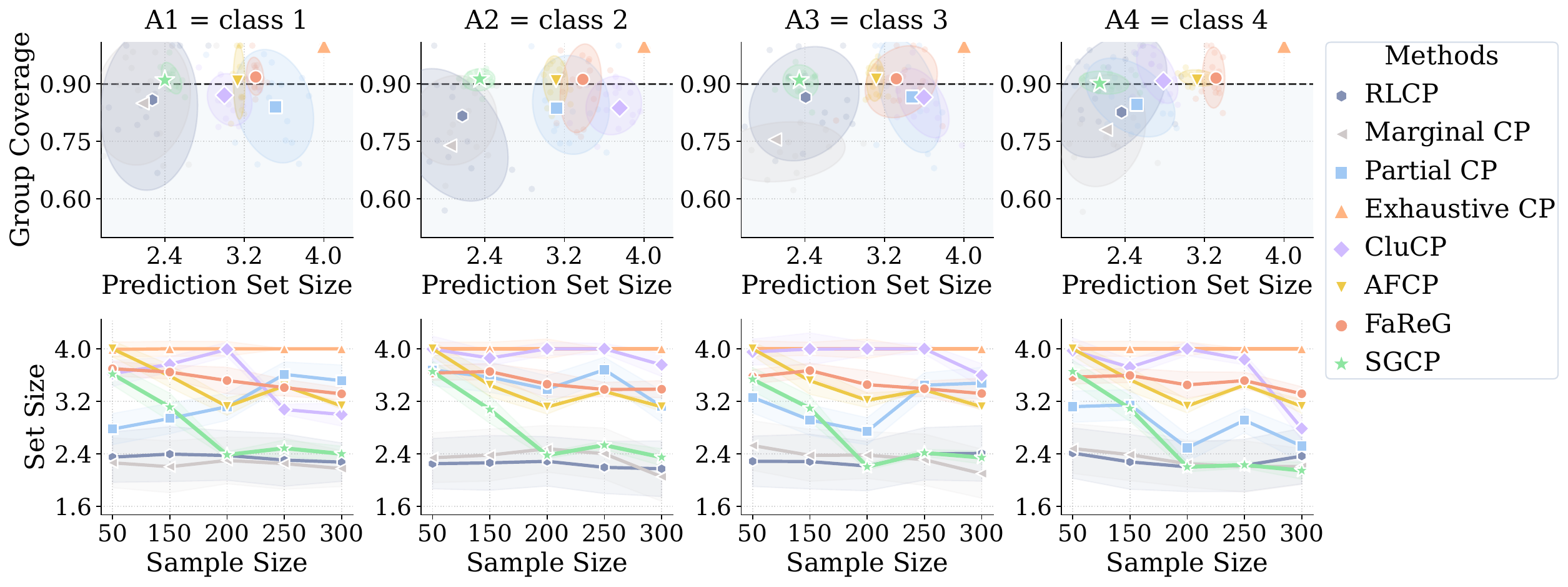}
    \caption{Group coverage on BACH dataset. \modelname remains close to the target coverage across classes while using smaller prediction sets than more conservative baselines.}
    \label{fig_bach_group}
\end{figure}

\begin{figure}
    \centering
    \includegraphics[width=0.95\linewidth]{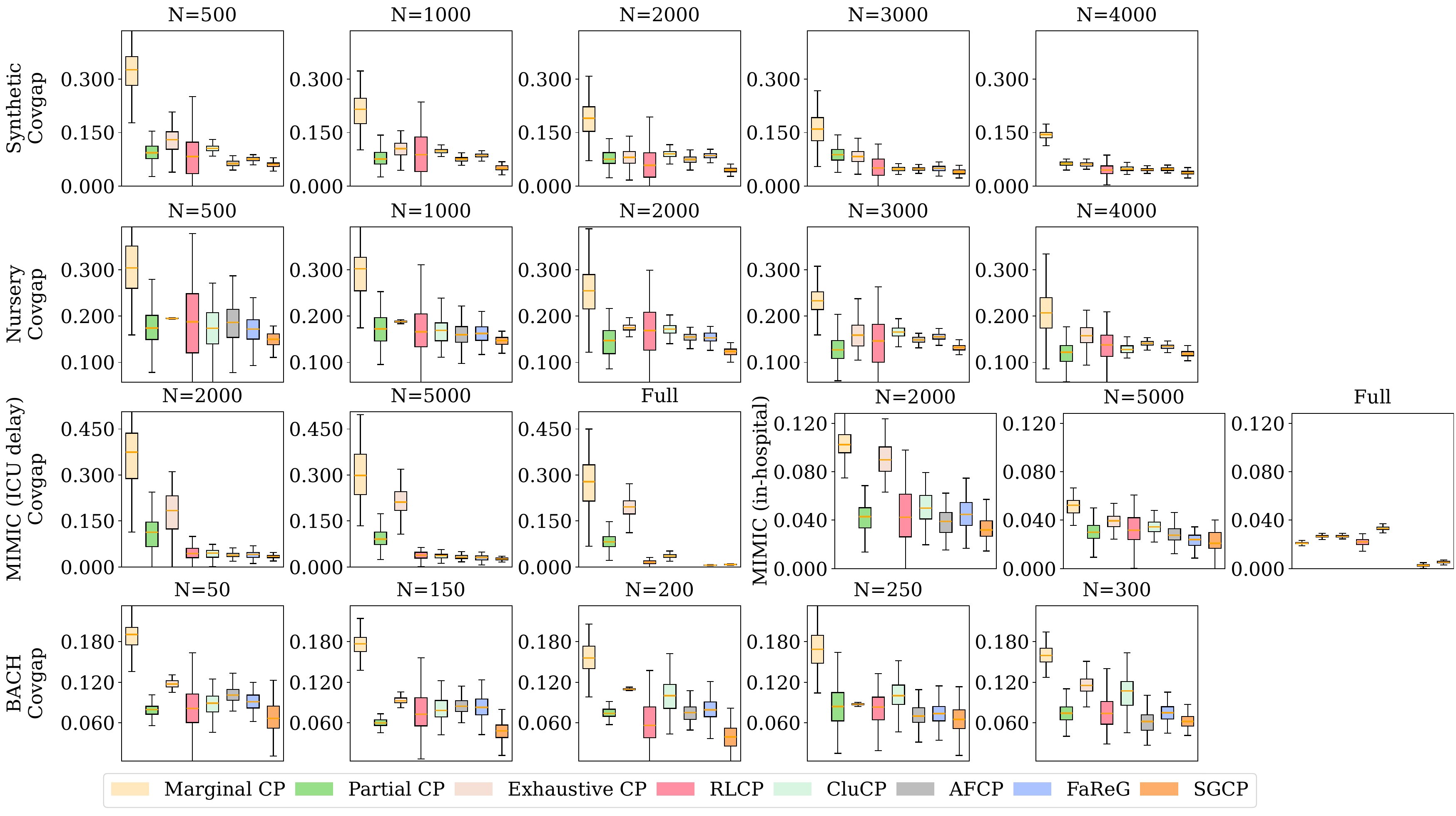}
    \caption{Distribution of CovGap over repeated runs under varying sample sparsity.
    }
    \vspace{-2mm}
    \label{fig_covgap_hist}
\end{figure}

\textbf{Main results.}
Table~\ref{tab_main_results} compares all methods across the four benchmarks (other experiments cf. Appendix~\ref{appendix_exp}).
AFCP is omitted on MIMIC-IV because its adaptive group selection requires an expensive repeated calibration procedure, making 20-run evaluation computationally prohibitive.
We highlight AvgSize values exceeding half of the label space in red, as prediction sets of this size can reduce clinical utility, increase the cognitive burden on decision makers.

We highlight four main observations.
\textbf{(i)} \modelname is the only method that consistently stays close to nominal coverage across all benchmarks while achieving strong subgroup reliability. 
Its empirical coverage remains near $0.9$, and its CovGap is the smallest on SYN, Nursery, and BACH, with values $4.2$, $11.8$, and $6.2$, respectively. 
On MIMIC-IV, it is close to the best result, with CovGap $0.8$ compared with FaReG's $0.5$.
\textbf{(ii)} The subgroup reliability gains of \modelname are not obtained through indiscriminate conservatism. \modelname achieves the smallest or nearly smallest AvgSize across all datasets. Taken together, these results show that \modelname improves subgroup reliability without enlarging prediction sets.
\textbf{(iii)} Existing baselines exhibit less favorable or less consistent tradeoffs.  
Marginal CP often leaves large subgroup disparities, while explicit subgroup-protection methods reduce disparity at the cost of inflated coverage and larger prediction sets. 
RLCP, CluCP, AFCP, and FaReG are competitive in some settings, but their gains are less uniform across benchmarks.

Figure~\ref{fig_pdf} visualizes the latent component densities learned by \modelname.
Across datasets, the components exhibit diverse shapes in the nonconformity score space.
Some components concentrate near small scores, while others are broader or place more mass on larger scores.
This suggests that \modelname learns nontrivial score law structure, rather than merely reproducing a single global score distribution.
Importantly, these components should not be viewed as demographic groups or semantic subpopulations.
They are latent score law building blocks.
For each sample, \modelname mixes these shared components to form a sample dependent local score law.
In this way, samples with similar calibration behavior can share evidence even when they do not belong to the same observed subgroup.
This provides a mechanism for improving subgroup reliability without requiring calibration to be tied to a single observed group label.



\textbf{Group coverage.}
To better understand where the improvement comes from, we visualize subgroup coverage against average prediction-set size, together with how set size varies with subgroup sample size.
Figure~\ref{fig_bach_group} corresponds to BACH, where the subgroup variable is the class label.
Two patterns are clear. First, \modelname remains close to the nominal target coverage while lying on the more efficient side of the tradeoff: in the top row, its points stay near the horizontal reference line at $0.9$ and are typically located to the left of competing methods.
Second, in the bottom row, its prediction sets shrink steadily as subgroup sample size increases, indicating that it can use additional subgroup evidence efficiently while still adapting locally.
Overall, these subgroup-level plots show that the gains of \modelname are realized across individual groups rather than only in aggregate.

\textbf{Distribution of CovGap.}
Figure~\ref{fig_covgap_hist} shows the distribution of CovGap over repeated runs under varying levels of sample sparsity. Beyond low average disparity, \modelname also exhibits stability: across datasets and sample sizes, it is consistently among the methods with the lowest median CovGap and relatively small variation across runs. This advantage is most pronounced in sparse-data settings, where subgroup calibration is most difficult.
As sample size grows, the gap narrows for most methods, but \modelname remains competitive and typically retains one of the lowest CovGap values.

\section{Conclusion}
In this work, we discussed the limitation of standard split-conformal prediction in fairness-sensitive settings, where marginal validity can mask substantial subgroup miscoverage. 
We argued that subgroup reliability is closely tied to the local distribution of nonconformity scores, and that  modeling this local score law provides a principled way to improve calibration across heterogeneous subpopulations. 
To this end, we proposed \modelname, a fairness-aware conformal prediction method that improves coverage for clinically meaningful subgroups without relying on predefined sensitive groups.
Our theoretical analysis showed that \modelname preserves the standard split-conformal marginal coverage guarantee and accurate local score-law learning reduces subgroup coverage gaps by making calibration behavior more comparable across subpopulations.
Extensive experiments across synthetic, and real-world benchmarks demonstrated that \modelname consistently improves subgroup reliability while maintaining efficient prediction sets. 
In future work, we plan to extend \modelname to personalized medicine, where reliable uncertainty quantification across diverse patient populations is crucial.

\section*{Acknowledgments}
This work was supported by the National Natural Science Foundation of China (Grant No. 62572097). We gratefully acknowledge the support provided by the Center for HPC at the University of Electronic Science and Technology of China.

\bibliographystyle{abbrv}
\bibliography{ref}

\appendix
\setcounter{figure}{5}
\setcounter{table}{1}
\setcounter{equation}{9}
\renewcommand{\theequation}{A\arabic{equation}}
\renewcommand{\thefigure}{A\arabic{figure}}
\renewcommand{\thetable}{A\arabic{table}}

\newpage

\section{Mathematical Proofs}

\subsection{Proof of Proposition~\ref{prop_oracle}}\label{appendix_proof_scp}

\begin{proof}
Let $S:=s(x,y)$ and $V:=\widehat F(S\mid X)$. 
Let $\lambda$ denote the Lebesgue measure on $[0,1]$. 
By the oracle assumption, for $P_X$-a.e. $x$,
\begin{equation*}
    \widehat F(\cdot\mid x)=F_x^\star(\cdot),
    \qquad
    F_x^\star(t):=\mathbb P(S\le t\mid X=x).
\end{equation*}
Moreover, by assumption, the conditional distribution of $S\mid X=x$ is continuous for $P_X$-a.e. $x$. Hence, by the probability integral transform, for $P_X$-a.e. $x$,
\begin{equation*}
    F_x^\star(S)\mid X=x \sim \mathrm{Unif}(0,1).
\end{equation*}
Since $V=\widehat F(S\mid X)=F_X^\star(S)$ almost surely, it follows that, for every Borel set $A\subseteq[0,1]$,
\begin{equation*}
    \mathbb E\!\left[\mathbf 1_{\{V\in A\}}\mid X\right]
    =
    \lambda(A)
    \qquad \text{a.s.}
\end{equation*}

Now let $G\subseteq\mathcal X$ be any measurable subgroup with
$\mathbb P(X\in G)>0$. Using the tower property,
\begin{equation*}
\begin{aligned}
\mathbb P(V\in A\mid X\in G)
&=
\frac{
    \mathbb E\!\left[
        \mathbf 1_{\{V\in A\}}\mathbf 1_{\{X\in G\}}
    \right]
}{
    \mathbb P(X\in G)
} \\ &=
\frac{
    \mathbb E\!\left[
        \mathbf 1_{\{X\in G\}}
        \mathbb E\!\left[\mathbf 1_{\{V\in A\}}\mid X\right]
    \right]
}{
    \mathbb P(X\in G)
} \\ &=
\frac{
    \mathbb E\!\left[
        \mathbf 1_{\{X\in G\}}\lambda(A)
    \right]
}{
    \mathbb P(X\in G)
} \\[0.5em]
&=
\lambda(A).
\end{aligned}
\end{equation*}
Therefore, for every set $A\subseteq[0,1]$, $\mathbb P(V\in A\mid X\in G)=\lambda(A)$, which is precisely the statement that
\begin{equation*}
    V\mid\{X\in G\}\sim \mathrm{Unif}(0,1).
\end{equation*}
\end{proof}

\begin{remark}
When this oracle local score law is recovered, the local percentile transform places scores from different inputs and subgroups on a common uniform scale, thereby eliminating subgroup reliability inconsistency. 
The objective of \modelname is to approximate this. Put differently, this proposition clarifies why we learn $\widehat{F}(s\mid x)$ instead of operating directly on the raw score $s$.
\end{remark}

\subsection{Proof of Proposition~\ref{prop_validity}}\label{proof_validity}
Throughout the proof, we condition on the training split $\mathcal D_{\mathrm{tra}}$. Thus the fitted transformation $(x,y)\mapsto \widehat F(s(x,y)\mid x)$ is fixed. For simplicity, we suppress this conditioning from the notation.

Let $m:=|\mathcal D_{\mathrm{cal}}|$ and write $\mathcal D_{\mathrm{cal}}=\{(X_i,Y_i)\}_{i=1}^m$
Define the transformed calibration scores
\[
V_i:=\widehat F(s(X_i,Y_i)\mid X_i),
\qquad i=1,\dots,m,
\]
and the transformed score of the test point under its true label
\[
V_\mathrm{test}:=\widehat F(s(X_{n+1},Y_{n+1})\mid X_{n+1}).
\]
Let $V_{(1)}\le \cdots \le V_{(m)}$ denote the order statistics of the calibration scores, and set
\[
k:=\left\lceil (m+1)(1-\alpha)\right\rceil .
\]
Define the split-conformal cutoff by $\widehat q_{1-\alpha}=V_{(k)}$, with the convention $\widehat q_{1-\alpha}=1$ whenever $k>m$.

\begin{proof}
By assumption, conditional on $\mathcal D_{\mathrm{tra}}$, the calibration examples in $\mathcal D_{\mathrm{cal}}$ and the test point $(X_{n+1},Y_{n+1})$ are exchangeable.
Since the same deterministic measurable map is applied to each point, it follows that
$V_1,\dots,V_m,V_\mathrm{test}$ are exchangeable conditional on $\mathcal D_{\mathrm{tra}}$.

Let $V_{(1)}\le \cdots \le V_{(m}$ denote the ordered transformed calibration scores, and define the split conformal cutoff by $\widehat q_{1-\alpha}
= V_{\left(\lceil (m+1)(1-\alpha)\rceil\right)}$, with the convention $\widehat q_{1-\alpha}=1$ whenever $\lceil (m+1)(1-\alpha)\rceil>m$.
By exchangeability and the standard split conformal order-statistic argument, we have $ \mathbb P \left(
V_\mathrm{test}\le \widehat q_{1-\alpha}
\;\middle|\;
\mathcal D_{\mathrm{tra}}
\right)
\ge 1-\alpha.
$

It remains to connect this event to the prediction set. By definition,
\begin{equation*}
    \widehat C_\alpha(X_{n+1})
=
\left\{
y\in\mathcal Y:\;
\widehat F\bigl(s(X_{n+1},y)\mid X_{n+1}\bigr)\le \widehat q_{1-\alpha}
\right\}.
\end{equation*}
Substituting the true label $Y_{n+1}$ gives
\begin{equation*}
\left\{
Y_{n+1}\in \widehat C_\alpha(X_{n+1})
\right\}
=\left\{
\widehat F\bigl(s(X_{n+1},Y_{n+1})\mid X_{n+1}\bigr)\le \widehat q_{1-\alpha}
\right\}
=
\left\{
V_\mathrm{test}\le \widehat q_{1-\alpha}
\right\}.
\end{equation*}

Hence,
\begin{equation*}
        \mathbb P\!\left(
        Y_{n+1}\in \widehat C_\alpha(X_{n+1})
        \,\middle|\,
        \mathcal D_{\mathrm{tra}}
    \right)
    \ge 1-\alpha.
\end{equation*}
Taking expectations over $\mathcal D_{\mathrm{tra}}$ gives
\begin{equation*}
    \mathbb P \left\{
    Y_{n+1}\in \widehat C_\alpha(X_{n+1})
    \right\}
    =
    \mathbb E \left[
    \mathbb P \left\{
    Y_{n+1}\in \widehat C_\alpha(X_{n+1})
    \,\middle|\, \mathcal D_{\mathrm{tra}}
    \right\}
    \right] 
    \ge 1-\alpha.
\end{equation*}
\end{proof}

\begin{remark}
In the current implementation, the stochastic grouping map $w(x)$ may be approximated by Monte Carlo averaging over latent posterior samples. The validity proof above continues to hold as long as the transformed score is computed using a rule that is fixed conditional on $\mathcal D_{\mathrm{tra}}$, or, more generally, uses auxiliary randomness that is independent across calibration and test samples.
\end{remark}

\subsection{Proof of Proposition~\ref{prop_approx_group_coverage}}\label{proof_approx_group_coverage}
\begin{proof}
By definition of the prediction set,
\[
\widehat C_\alpha(X_{n+1})
=
\left\{
y\in\mathcal Y:
\widehat F(s(X_{n+1},y)\mid X_{n+1})
\le
\widehat q_{1-\alpha}
\right\}.
\]
Therefore,
\[
\left\{
Y_{n+1}\in \widehat C_\alpha(X_{n+1})
\right\}
=
\left\{
V_\mathrm{test}\le \widehat q_{1-\alpha}
\right\}.
\]

Now fix a measurable subgroup $G\subseteq\mathcal X$ with $\mathbb P(X\in G)>0$.
Since the calibration sample is independent of the test point, $\widehat q_{1-\alpha}$ is independent of $(X_{n+1},Y_{n+1})$.
Hence, conditional on $\widehat q_{1-\alpha}$,
\[
\begin{aligned}
&\mathbb P\!\left(
Y_{n+1}\in \widehat C_\alpha(X_{n+1})
\,\middle|\,
X_{n+1}\in G,\widehat q_{1-\alpha}
\right) \\
&\qquad =
\mathbb P\!\left(
V_{n+1}\le \widehat q_{1-\alpha}
\,\middle|\,
X_{n+1}\in G,\widehat q_{1-\alpha}
\right) \\
&\qquad =
H_G(\widehat q_{1-\alpha}),
\end{aligned}
\]
where
\[
H_G(t):=\mathbb P(V(X,Y)\le t\mid X\in G).
\]
Taking expectations over $\widehat q_{1-\alpha}$ gives
\[
\mathbb P\!\left(
Y_{n+1}\in \widehat C_\alpha(X_{n+1})
\,\middle|\,
X_{n+1}\in G
\right)
=
\mathbb E\!\left[
H_G(\widehat q_{1-\alpha})
\right].
\]

Next, let
\[
H(t):=\mathbb P(V(X,Y)\le t)
\]
be the global CDF of the transformed score, and let $V^\dagger$ be an independent draw from this distribution.
Since $V_1,\dots,V_m,V^\dagger$ are exchangeable and $H$ is continuous, the standard split-conformal rank argument gives
\[
\mathbb P\!\left(
V^\dagger\le \widehat q_{1-\alpha}
\right)
=
\frac{k}{m+1}.
\]
On the other hand, conditioning on $\widehat q_{1-\alpha}$,
\[
\mathbb P\!\left(
V^\dagger\le \widehat q_{1-\alpha}
\,\middle|\,
\widehat q_{1-\alpha}
\right)
=
H(\widehat q_{1-\alpha}).
\]
Therefore,
\[
\mathbb E\!\left[
H(\widehat q_{1-\alpha})
\right]
=
\frac{k}{m+1}.
\]

Combining the two identities,
\[
\begin{aligned}
&\left|
\mathbb P\!\left(
Y_{n+1}\in \widehat C_\alpha(X_{n+1})
\,\middle|\,
X_{n+1}\in G
\right)
-
\frac{k}{m+1}
\right| \\
&\qquad =
\left|
\mathbb E\!\left[
H_G(\widehat q_{1-\alpha})
\right]
-
\mathbb E\!\left[
H(\widehat q_{1-\alpha})
\right]
\right| \\
&\qquad =
\left|
\mathbb E\!\left[
H_G(\widehat q_{1-\alpha})
-
H(\widehat q_{1-\alpha})
\right]
\right| \\
&\qquad \le
\mathbb E\!\left[
\left|
H_G(\widehat q_{1-\alpha})
-
H(\widehat q_{1-\alpha})
\right|
\right] \\
&\qquad \le
\sup_{t\in[0,1]}
\left|
H_G(t)-H(t)
\right|
=
\Delta_G.
\end{aligned}
\]
This proves the claimed subgroup coverage gap bound.

Finally, since
\[
\frac{k}{m+1}
=
\frac{\lceil (m+1)(1-\alpha)\rceil}{m+1}
\ge
1-\alpha,
\]
we obtain
\[
\mathbb P\!\left(
Y_{n+1}\in \widehat C_\alpha(X_{n+1})
\,\middle|\,
X_{n+1}\in G
\right)
\ge
1-\alpha-\Delta_G.
\]
\end{proof}

Since $k/(m+1)\ge 1-\alpha$, this result implies that whenever a subgroup's transformed-score distribution is close to the overall transformed-score distribution, the global conformal cutoff keeps its coverage close to the nominal level $1-\alpha$.

\begin{remark}
Proposition~\ref{prop_approx_group_coverage} bounds subgroup miscoverage in terms of the residual transformed-score mismatch $\Delta_G$.
This mismatch can be further related to local score-law estimation error.
If the learned transform satisfies
\[
\sup_{x,s}
|\widehat F(s\mid x)-F_x^\star(s)|
\le \varepsilon,
\]
then, by the probability integral transform,
\[
\Delta_G\le 2\varepsilon
\]
for every measurable subgroup $G$.
Consequently,
\[
\left|
\mathbb P\{Y_{n+1}\in\widehat C_\alpha(X_{n+1})\mid X_{n+1}\in G\}
-
(1-\alpha)
\right|
\le
2\varepsilon+\frac{1}{m+1}.
\]
Thus, accurate local score-law learning provides a sufficient condition for small subgroup coverage gaps.
\end{remark}

\subsection{Empirical Subgroup Coverage}\label{cor_lowerbound}
We now relate the empirical subgroup coverage measured on an independent test sample to the population subgroup coverage above.
Under the assumptions of Proposition~\ref{prop_approx_group_coverage}, let
$
\mathcal D_G^{\mathrm{test}}
=
\{(X_j^{(G)},Y_j^{(G)})\}_{j=1}^{n_G}
$
be an i.i.d. test sample drawn from the conditional distribution of $(X,Y)$ given $X\in G$, independent of $\mathcal D_{\mathrm{tra}}$ and $\mathcal D_{\mathrm{cal}}$. 

\begin{corollary}
\label{cor_empirical_subgroup_gap}
Let $m=|\mathcal D_{\mathrm{cal}}|$ and $k=\left\lceil(m+1)(1-\alpha)\right\rceil$, and assume $k\leq m$. Let $\widehat{\mathrm{Cov}}_G$ be the coverage measured on an independent subgroup test sample of size $n_G$, denoted as
\begin{equation*}
\widehat{\mathrm{Cov}}_G:=\frac{1}{n_G}\sum_{j=1}^{n_G}\mathbf 1\left\{Y_j^{(G)}\in\widehat C_\alpha(X_j^{(G)})\right\}.
\end{equation*}
Then, for any $\delta\in(0,1)$, conditional on $\mathcal D_{\mathrm{tra}}$, with probability at least $1-\delta$ jointly over $\mathcal D_{\mathrm{cal}}$ and $\mathcal D_G^{\mathrm{test}}$,
\begin{equation*}
\left|\widehat{\mathrm{Cov}}_G-(1-\alpha)\right|
\leq
\Delta_G+\sqrt{\frac{\log(4/\delta)}{2n_G}}+\sqrt{\frac{\log(4/\delta)}{2m}}+\left|\frac{k}{m}-(1-\alpha)\right|.
\end{equation*}
In particular,
\begin{equation*}
\widehat{\mathrm{Cov}}_G
\geq
1-\alpha-\Delta_G-\sqrt{\frac{\log(4/\delta)}{2n_G}}-\sqrt{\frac{\log(4/\delta)}{2m}}.
\end{equation*}
\end{corollary}

Corollary~\ref{cor_empirical_subgroup_gap} connects the theoretical subgroup discrepancy to the coverage measured on an independent finite subgroup test sample. When the learned transform reduces $\Delta_G$ and both the calibration sample size $m$ and subgroup test sample size $n_G$ are sufficiently large, the empirical subgroup coverage concentrates near the nominal level $1-\alpha$. The two concentration terms account for uncertainty in the calibration cutoff and sampling variation in the subgroup test set, while the final deterministic term accounts for quantile discretization.

\begin{proof}
Throughout the proof, we condition on $\mathcal D_{\mathrm{tra}}$, so that the learned score transformation is fixed. Let $V_1,\dots,V_m$ be the transformed calibration scores, let
\begin{equation*}
\widehat H_m(t):=\frac{1}{m}\sum_{i=1}^{m}\mathbf 1\left\{V_i\leq t\right\},
\end{equation*}
and let $\widehat q_{1-\alpha}=V_{(k)}$. By continuity of the marginal transformed-score distribution, the calibration scores have no ties almost surely, and hence
\begin{equation}
\widehat H_m(\widehat q_{1-\alpha})=\frac{k}{m}.
\label{eq_empirical_cdf_at_cutoff}
\end{equation}

Let $H$ and $H_G$ denote the marginal and subgroup CDFs of the transformed score, respectively, conditional on $\mathcal D_{\mathrm{tra}}$. By definition,
\begin{equation*}
\Delta_G=\sup_{t\in[0,1]}\left|H_G(t)-H(t)\right|.
\end{equation*}
The Dvoretzky--Kiefer--Wolfowitz inequality gives
\begin{equation}
\mathbb P\left(\sup_{t\in[0,1]}\left|\widehat H_m(t)-H(t)\right|\geq\sqrt{\frac{\log(4/\delta)}{2m}}\right)\leq\frac{\delta}{2}.
\label{eq_dkw_calibration}
\end{equation}

Conditional on $\mathcal D_{\mathrm{tra}}$ and $\mathcal D_{\mathrm{cal}}$, the prediction set $\widehat C_\alpha$ is fixed. Since the test points
\begin{equation*}
(X_j^{(G)},Y_j^{(G)}),\qquad j=1,\dots,n_G,
\end{equation*}
are i.i.d. draws from the conditional distribution of $(X,Y)$ given $X\in G$, the indicators
\begin{equation*}
Z_j:=\mathbf 1\left\{Y_j^{(G)}\in\widehat C_\alpha(X_j^{(G)})\right\},\qquad j=1,\dots,n_G,
\end{equation*}
are i.i.d. Bernoulli random variables with mean
\begin{equation*}
\mathrm{Cov}_G(\mathcal D_{\mathrm{cal}}):=\mathbb P\left\{Y\in\widehat C_\alpha(X)\right\}=H_G(\widehat q_{1-\alpha}).
\end{equation*}
Therefore, by Hoeffding's inequality,
\begin{equation}
\mathbb P\left(\left|\widehat{\mathrm{Cov}}_G-\mathrm{Cov}_G(\mathcal D_{\mathrm{cal}})\right|\geq\sqrt{\frac{\log(4/\delta)}{2n_G}}\right)\leq\frac{\delta}{2}.
\label{eq_hoeffding_subgroup}
\end{equation}

By Eqs.~\eqref{eq_dkw_calibration} and \eqref{eq_hoeffding_subgroup} and a union bound, both events hold jointly with probability at least $1-\delta$. On this event,
\begin{align*}
\left|\widehat{\mathrm{Cov}}_G-(1-\alpha)\right|
&\leq \left|\widehat{\mathrm{Cov}}_G-\mathrm{Cov}_G(\mathcal D_{\mathrm{cal}})\right|
+\left|H_G(\widehat q_{1-\alpha})-H(\widehat q_{1-\alpha})\right| \\
&\quad+\left|H(\widehat q_{1-\alpha})-\widehat H_m(\widehat q_{1-\alpha})\right|
+\left|\widehat H_m(\widehat q_{1-\alpha})-(1-\alpha)\right| \\
&\leq \sqrt{\frac{\log(4/\delta)}{2n_G}}+\Delta_G+\sqrt{\frac{\log(4/\delta)}{2m}}+\left|\frac{k}{m}-(1-\alpha)\right|,
\end{align*}
where the last inequality uses Eq.~\eqref{eq_empirical_cdf_at_cutoff}.

Finally, on the same high-probability event,
\begin{align*}
\widehat{\mathrm{Cov}}_G
&\geq \mathrm{Cov}_G(\mathcal D_{\mathrm{cal}})-\sqrt{\frac{\log(4/\delta)}{2n_G}} \\
&\geq H(\widehat q_{1-\alpha})-\Delta_G-\sqrt{\frac{\log(4/\delta)}{2n_G}} \\
&\geq \frac{k}{m}-\Delta_G-\sqrt{\frac{\log(4/\delta)}{2n_G}}-\sqrt{\frac{\log(4/\delta)}{2m}} \\
&\geq 1-\alpha-\Delta_G-\sqrt{\frac{\log(4/\delta)}{2n_G}}-\sqrt{\frac{\log(4/\delta)}{2m}},
\end{align*}
where the last inequality follows from $k/m\geq1-\alpha$.
\end{proof}

\section{Model Details}
\label{app_implementation_sgcp}
This appendix specifies the concrete instantiation of the aggregate local score law in Eq.~\eqref{eq_local_score_law} and the corresponding training objective.
The algorithm~\ref{alg_detailed} demonstrates a specific process.

\paragraph{Latent calibration state.}
Let $h_\theta(x)\in\mathbb R^d$ denote the backbone feature representation and let $f_\theta(x)\in\Delta^{C-1}$ denote the corresponding predictive probability vector.
We parameterize a Gaussian latent posterior
$q_\phi(z\mid x)
=
\mathcal N \Bigl(
\mu_\phi(h_\theta(x)),
\operatorname{diag}(\sigma_\phi^2(h_\theta(x)))
\Bigr)$.

\begin{algorithm}[t]
\caption{Detailed Training and Inference Procedure for \modelname}
\label{alg_detailed}
\begin{algorithmic}[1]
\Require
Training data $\mathcal D_{\mathrm{tra}}$, calibration data $\mathcal D_{\mathrm{cal}}$, backbone classifier $f_\theta$, nonconformity score $s(\cdot,\cdot)$, $K$ latent calibration components, $M$ posterior samples, hyperparameters $\lambda, \beta$, miscoverage level $\alpha$
\Ensure
Prediction set $\widehat C_\alpha(X_{N+1})$

\State Split $\mathcal D_{\mathrm{tra}}$ into $\mathcal D_0$ and $\mathcal D_1$
\State Train the backbone predictor $f_\theta$ on $\mathcal D_0$
\State Initialize the stochastic assignment module and component laws $\{F_k\}_{k=1}^K$

\For{each optimization step on $\mathcal D_1$}
    \State Sample a mini-batch $\mathcal B=\{(X_i,Y_i)\}_{i=1}^B$
    \State Compute backbone features, and nonconformity scores $s_i=s(X_i,Y_i)$
    \State Estimate posterior-averaged memberships from $q_\phi(z\mid X_i)$
    \State Compute component transforms $C_{ik}=F_k(S_i)$ and local densities $\widehat f(S_i\mid X_i)=\sum_{k=1}^K w_{ik}f_k(S_i)$
    \State Compute $\mathcal L_{\mathrm{rank}}$, $\mathcal L_{\mathrm{nll}}$, and $\widehat I_{\mathcal B}^{\,\mathrm{norm}}(X;Z)$
    \State Update assignment and component parameters by minimizing $\mathcal L_{SGCP}$
\EndFor

\State Construct the learned local score transform $\widehat F(s\mid x)=\sum_{k=1}^K w_k(x)F_k(s)$

\For{each $(X_i,Y_i)\in\mathcal D_{\mathrm{cal}}$}
    \State Compute $V_i=\widehat F(s(X_i,Y_i)\mid X_i)$
\EndFor

\State Sort $\{V_i\}_{i=1}^m$ and compute the conformal cutoff $\widehat q_{1-\alpha}=V_{(\lceil (m+1)(1-\alpha)\rceil)}$ with fallback $\widehat q_{1-\alpha}=1$ if $\lceil (m+1)(1-\alpha)\rceil>m$

\For{each candidate label $y\in\mathcal Y$}
    \State Compute
    $V_\mathrm{test}^y=\widehat F(S(X_{n+1},y)\mid X_{N+1})$
\EndFor

\State \Return
$\widehat C_\alpha(X_{n+1})
=
\{y\in\mathcal Y:V_{n+1}^y\le \widehat q_{1-\alpha}\}$

\end{algorithmic}
\end{algorithm}

\paragraph{Stochastic grouping map.}\label{appendx_consistency}
Let $c_1,\dots,c_K\in\mathbb R^r$ denote $K$ learnable prototypes in the latent space. For a latent draw $z\sim q_\phi(z\mid x)$, define the assignment vector
\[
\tilde w_k(x,z)
=
\frac{
\exp(-\|z-c_k\|_2^2/T)
}{
\sum_{j=1}^K \exp(-\|z-c_j\|_2^2/T)
},
\qquad k=1,\dots,K,
\]
where $T>0$ is a temperature parameter. 
Given $z^{(1)},\dots,z^{(M)}\overset{\mathrm{i.i.d.}}{\sim}q_\phi(z\mid x)$, the posterior-averaged membership vector
$w(x)=(w_1(x),\dots,w_K(x))\in\Delta^{K-1}$ is estimated by
\[
w_k(x)
=
\mathbb E_{z\sim q_\phi(\cdot\mid x)}[\tilde w_k(x,z)]
\approx
\frac{1}{M}\sum_{m=1}^M \tilde w_k(x,z^{(m)}),
\qquad k=1,\dots,K.
\]

\begin{remark}
Marginal conformal validity only requires global exchangeability, so it can hold even when score behavior varies substantially across subgroups. 
Such heterogeneity can make a single global calibration rule locally unreliable. 
\modelname addresses this by learning latent calibration components that shift calibration evidence among samples with similar score behavior, providing a practical way to approximate oracle local score laws without assuming exact local exchangeability.
\end{remark}

\begin{remark}
In SGCP, posterior sampling approximates a stable stochastic grouping map rather than adding randomness to conformal calibration. 
For fixed $x$, draw $z^{(1)},\dots,z^{(M)}\overset{\mathrm{i.i.d.}}{\sim}q_\phi(z\mid x)$ and set
\[
\widehat w_k^{(M)}(x)
=
\frac{1}{M}\sum_{m=1}^M \tilde w_k(x,z^{(m)}),
\qquad
w_k(x)=\mathbb E_{z\sim q_\phi(\cdot\mid x)}[\tilde w_k(x,z)].
\]
The induced local score law is $\widehat F^{(M)}(s\mid x)=
\sum_{k=1}^K \widehat w_k^{(M)}(x)F_k(s)$.

Samples share calibration evidence through the common components $\{F_k\}_{k=1}^K$, and posterior averaging reduces Monte Carlo variability in the sample-specific weights. 
As $M\to\infty$, the law of large numbers gives $\widehat w_k^{(M)}(x)\to w_k(x)$ almost surely, and therefore
\[
\widehat F^{(M)}(s\mid x)
\to
\widehat F(s\mid x)
=
\sum_{k=1}^K w_k(x)F_k(s)
\quad \text{a.s.}
\]
\end{remark}

\subsection{Training Objective}

We train the stochastic assignment module and the component parameters on $\mathcal D_1$, while keeping the backbone fixed.
For each $(X_i,Y_i)\in\mathcal D_1$, define
\[
S_i:=s(X_i,Y_i),
\qquad
w_{ik}:=w_k(X_i),
\qquad
C_{ik}:=F_k(S_i).
\]
The latent components are intended to serve as reusable local score laws.
Thus, scores routed to the same component should be comparable after being mapped to their component-wise percentile scale, while the induced local mixture density should still explain the observed true-label nonconformity scores.
These two requirements lead to the rank-uniformization and local density-fitting terms below.

\paragraph{Rank-uniformization.}
For a mini-batch $\mathcal B$ of size $B$ and a fixed component $k$, let
$\{C_{(1)k},\dots,C_{(B)k}\}$ be the sorted values of $\{C_{ik}:i\in\mathcal B\}$, and let
$w_{(1)k},\dots,w_{(B)k}$ denote the correspondingly reordered weights.
Normalize the weights within component $k$ by
\[
\pi_{(b)k}
=
\frac{w_{(b)k}}
{\sum_{\ell=1}^B w_{(\ell)k}} .
\]
The weighted rank midpoint is
$r_{(b)k}
=
\sum_{\ell<b}\pi_{(\ell)k}
+
\frac{1}{2}\pi_{(b)k}$.
We define the component-wise rank-uniformization loss as
\[
\ell_{\mathrm{rank}}^{(k)}
=
\sum_{b=1}^B
\pi_{(b)k}
\left(
C_{(b)k}-r_{(b)k}
\right)^2 .
\]
Averaging over active components gives
\[
\mathcal L_{\mathrm{rank}}
=
\frac{1}{|\mathcal K_{\mathrm{act}}|}
\sum_{k\in\mathcal K_{\mathrm{act}}}
\ell_{\mathrm{rank}}^{(k)},
\qquad
\mathcal K_{\mathrm{act}}
=
\left\{
k:\sum_{i\in\mathcal B} w_{ik}>\tau
\right\}.
\]

\paragraph{Local density fitting.}
Let $f_k$ denote the density associated with $F_k$.
The local mixture density at $(X_i,S_i)$ is $\widehat f(S_i\mid X_i)
= \sum_{k=1}^K w_{ik}f_k(S_i)$.
We fit the observed true-label scores by the negative log-likelihood
\[
\mathcal L_{\mathrm{nll}}
=
-\frac{1}{|\mathcal B|}
\sum_{i\in\mathcal B}
\log \widehat f(S_i\mid X_i).
\]
The rank term makes the component transforms behave like percentile maps, whereas the likelihood term ties the learned score laws to the empirical score distribution.

\paragraph{Information maximization for stochastic grouping.}
The preceding two terms specify how the component score laws should behave, but they do not by themselves ensure that the grouping map uses the components in an informative way.
For example, a balance constraint on the average component mass can prevent all samples from collapsing onto a single component, but it still permits the opposite degenerate solution in which every sample is assigned nearly uniformly to all components.
Such assignments are balanced in aggregate, yet they carry little sample-specific information and tend to make the learned component laws indistinguishable.

We therefore encourage the component assignment to retain information about the input.
For a mini-batch $\mathcal B$, view the assignment as a conditional categorical distribution $q_\phi(Z=k\mid X_i)=w_{ik}$, where $Z$ denotes the latent component index.
The induced marginal component distribution on the mini-batch is
\[
\bar w_k
=
\frac{1}{|\mathcal B|}
\sum_{i\in\mathcal B}w_{ik},
\qquad
\bar w=(\bar w_1,\dots,\bar w_K).
\]
The empirical mutual information between inputs and component assignments is
\[
\widehat I_{\mathcal B}(X;Z)
=
H(\bar w)
-
\frac{1}{|\mathcal B|}
\sum_{i\in\mathcal B} H(w_i),
\]
where
\[
H(\bar w)
=
-\sum_{k=1}^K \bar w_k\log \bar w_k,
\qquad
H(w_i)
=
-\sum_{k=1}^K w_{ik}\log w_{ik}.
\]
We use the normalized form
\[
\widehat I_{\mathcal B}^{\,\mathrm{norm}}(X;Z)
=
\frac{1}{\log K}
\left[
H(\bar w)
-
\frac{1}{|\mathcal B|}
\sum_{i\in\mathcal B} H(w_i)
\right].
\]
The marginal entropy term encourages all components to be used at the batch level, while the conditional entropy term encourages each sample to have a sharper, more informative assignment. 
Thus, maximizing $\widehat I_{\mathcal B}^{\,\mathrm{norm}}(X;Z)$ discourages both hard component collapse and soft uniform assignment.

\paragraph{Full objective.}
Taking the rank-uniformization loss as the reference scale, we optimize
\[
\mathcal L_{\mathrm{\modelname}}
=
\mathcal L_{\mathrm{rank}}
+
\lambda\mathcal L_{\mathrm{nll}}
-
\beta
\widehat I_{\mathcal B}^{\,\mathrm{norm}}(X;Z),
\]
where $\lambda$ controls the strength of local density fitting and $\beta$ controls the strength of stochastic-grouping information maximization.

\section{Complexity Analysis}

Beyond backbone training, \modelname adds three sources of computation: stochastic grouping, local score-law fitting, and conformal calibration on transformed scores. 
Let $N$ denote the number of training samples used for learning the grouping map and local score laws, $N_{\mathrm{cal}}$ the calibration size, and $N_{\mathrm{test}}$ the number of test samples. 
Treating the number of epochs, posterior samples, latent components, latent dimension, classes, and batch size as constants, the training overhead is linear in $N$, up to an optional $\mathcal O(N\log N)$ sorting step for constructing empirical component CDFs.

Calibration requires computing one transformed score per calibration sample and selecting a conformal quantile, costing $\mathcal O(N_{\mathrm{cal}}\log N_{\mathrm{cal}})$ with sorting or $\mathcal O(N_{\mathrm{cal}})$ with linear-time selection. 
At test time, the grouping map is evaluated once per input and transformed scores are computed for all candidate labels, giving constant overhead per test sample and total cost $\mathcal O(N_{\mathrm{test}})$.

Overall, under the standard mini-batch implementation, the additional complexity beyond backbone training is $\mathcal O(N+N_{\mathrm{cal}}\log N_{\mathrm{cal}}+N_{\mathrm{test}})$, or $\mathcal O(N+N_{\mathrm{cal}}+N_{\mathrm{test}})$ with linear-time quantile selection. 

\section{Detailed Experimental Setup}
\subsection{Additional Benchmark Details}\label{appendix_experiment}

\begin{table}[ht]
\centering
\small
\setlength{\tabcolsep}{5pt}
\caption{Summary of datasets and grouping attributes used for evaluation.}
\begin{tabular}{l l c c l}
\hline
Dataset & Modality & \#Classes & \#Attrs & Sensitive / grouping attributes \\
\hline
SYN & tabular & 6 & 3
& $A_1=$ gender, $A_2=$ phenotype, $A_3=$ age \\
Nursery & tabular & 4 & 5
& $A_1=$ parents, $A_2=$ children, $A_3=$ finance, \\
& & & & $A_4=$ social, $A_5=$ health \\
MIMIC-IV & tabular & 2/3 & 3
& $A_1=$ minority, $A_2=$ gender, $A_3=$ public insurance \\
BACH & image & 4 & -
& class label (for class-wise reliability analysis) \\
\hline
\end{tabular}
\label{tab:datasets}
\end{table}

In this section, we describe the dataset preprocessing in detail. 
A summary is provided in Table~\ref{tab:datasets}. 

\paragraph{Synthetic.}\label{appendix_syn}
The synthetic benchmark is a controlled multiclass cardiopulmonary diagnosis task with six labels,
$Y\in\{1,\dots,6\}$, corresponding to Hypertension, Coronary Artery Disease, Heart Failure, Arrhythmia, Valvular Heart Disease, and Peripheral Artery Disease.

Each sample contains four grouping attributes,
$A=(A_{\mathrm{sex}},A_{\mathrm{pheno}},A_{\mathrm{age}},A_{\mathrm{region}})$,
and six non-sensitive physiological variables $Z=(Z_1,\dots,Z_6)$, where
$Z_j\overset{\mathrm{i.i.d.}}{\sim}\mathcal U(0,1)$.
We draw
\[
A_{\mathrm{sex}}\sim\mathrm{Bernoulli}(1/2),
\qquad
A_{\mathrm{pheno}}\sim\mathrm{Bernoulli}(1/2),
\]
with $A_{\mathrm{sex}}\in\{0,1\}$ denoting Female/Male and
$A_{\mathrm{pheno}}\in\{0,1\}$ denoting Type I/Type II.
Age is assigned cyclically over
$\{\mathrm{Pediatric},\mathrm{Adult},\mathrm{Geriatric}\}$,
and region is sampled uniformly from
$\{\mathrm{North},\mathrm{South},\mathrm{East},\mathrm{West}\}$.

To induce intersection-dependent difficulty, we define the disadvantaged subgroup through the XNOR relation between sex and phenotype,
\[
G_{\mathrm{bias}}=\{i:\;A_{\mathrm{sex},i}=A_{\mathrm{pheno},i}\},
\]
which includes Female--Type I and Male--Type II individuals.

Labels are generated from a latent multiclass mechanism based on $X_i=(A_i,Z_i)$.
We first construct a class-probability vector $\pi(X_i)\in\Delta^5$ over the six disease classes.
Outside ambiguous regions of the feature space, labels follow the dominant class implied by $\pi(X_i)$.
Inside ambiguous regions, we introduce subgroup-dependent uncertainty by mixing $\pi(X_i)$ with the uniform distribution $u=(1/6,\dots,1/6)$:
\[
Y_i \sim
\begin{cases}
\mathrm{Cat}\bigl((1-\delta_1)\pi(X_i)+\delta_1 u\bigr), & i\in G_{\mathrm{bias}}, \\[1mm]
\mathrm{Cat}\bigl((1-\delta_0)\pi(X_i)+\delta_0 u\bigr), & i\notin G_{\mathrm{bias}},
\end{cases}
\qquad \delta_1>\delta_0.
\]
The parameters $\delta_1$ and $\delta_0$ control label ambiguity for the disadvantaged subgroup and the remaining population, respectively. 
Thus, $G_{\mathrm{bias}}$ exhibits systematically higher conditional uncertainty while the global task structure remains unchanged, yielding a controlled setting for studying subgroup-dependent calibration difficulty.

\paragraph{Nursery.}
Nursery is a public multiclass tabular benchmark derived from a hierarchical decision problem for ranking nursery-school applications.
Following standard preprocessing, we remove the \texttt{recommend} class, which contains only two samples, and encode the remaining labels into four admission-priority levels,
$Y\in\{0,1,2,3\}$.

Each instance contains eight categorical covariates describing household background and application context, including parents' occupation, number of children, housing, financial standing, social conditions, and health status.
All categorical variables are label-encoded before training.
We use five covariates as candidate sensitive attributes,
\[
A=(A_{\mathrm{parents}},A_{\mathrm{children}},A_{\mathrm{finance}},A_{\mathrm{social}},A_{\mathrm{health}}),
\]
corresponding to parents' occupation, number of children, financial standing, social conditions, and health status.
These attributes are used only for subgroup construction and fairness evaluation.

To induce controlled subgroup disadvantage, we follow the intervention protocol of prior work~\cite{10.5555/3737916.3741370,xu2026fair}.
We define
\[
G_{\mathrm{nur}}=\{i:X_{\mathrm{parents},i}=1\},
\]
corresponding to a particular preprocessed category of the \texttt{parents} attribute.
We then modify only this subgroup by downsampling it to $10\%$ of its original size and injecting label uncertainty.
Specifically, labels are perturbed by independent noise
$\varepsilon_i\sim\mathcal U(-4,4)$, then rounded and clipped to $\{0,1,2,3\}$.
This creates a subgroup that is both underrepresented and noisier than the rest of the population, serving as the disadvantaged group in our fairness evaluation.

\paragraph{MIMIC-IV.}\label{appendix_mimic}
We construct our clinical benchmark from MIMIC-IV at the hospital-admission level. 
Each admission is linked with ICU-stay information and diagnosis-code counts, yielding tabular covariates from demographic, administrative, and utilization records. 
Age is computed from \texttt{anchor\_age}, adjusted by the difference between admission year and anchor year, and clipped to $[0,89]$. 
We also derive ICU-related features, including ICU length of stay, time from hospital admission to first ICU entry, and the number of diagnosis codes.

For subgroup evaluation, we define three binary grouping attributes. 
The race indicator \texttt{minority} equals $1$ if the recorded race string does not contain \texttt{WHITE}; \texttt{gender\_m} equals $1$ for male patients; and \texttt{public\_insurance} equals $1$ if the primary insurance is \texttt{MEDICARE} or \texttt{MEDICAID}. 
These attributes are used only for subgroup construction and fairness evaluation. 
Model inputs include age, one-hot indicators for insurance type, admission type, and marital status, ICU length of stay, and the number of diagnosis codes; ICU length of stay is excluded for ICU-delay prediction.

We consider two tasks. 
The first is in-hospital mortality prediction, with binary labels derived from \texttt{hospital\_expire\_flag}. 
Results for this task are reported as MIMIC-IV (in-hospital) in Figures~\ref{fig_covgap_hist}, \ref{fig_coverage_hist}, and~\ref{fig_size_hist}. 
The second is ICU-delay prediction, reported as MIMIC-IV (ICU delay), where admissions are categorized by the time from hospital admission to first ICU entry:
\[
Y =
\begin{cases}
0, & \text{if ICU delay } \leq 4 \text{ hours},\\
1, & \text{if ICU delay } \in (4,12] \text{ hours},\\
2, & \text{if ICU delay } > 12 \text{ hours or no ICU admission occurs}.
\end{cases}
\]
Together, these tasks provide a real-world clinical benchmark for evaluating fairness across race, gender, and insurance-defined subgroups.

\begin{table*}[t]
\centering
\caption{Dataset specific hyperparameter settings for \modelname.}
\label{tab_hparams}
\resizebox{\linewidth}{!}{
\begin{tabular}{lcccccc}
\toprule
Hyperparameter & SYN & Nursery & M-ICU & M-Hosp & BACH & Description \\
\midrule

\multicolumn{7}{l}{\textit{\textbf{(i) Latent calibration components}}} \\
$K$ & 3 & 6 & 3 & 3 & 6 & number of latent calibration components \\
$d_z$ & 16 & 8 & 8 & 8 & 64 & latent calibration dimension \\
hidden dim & 64 & 64 & 32 & 32 & 64 & hidden dimension of grouping network \\

\addlinespace[0.5em]
\multicolumn{7}{l}{\textit{\textbf{(ii) Score law learning}}} \\
\modelname train epochs & 100 & 100 & 100 & 100 & 100 & number of \modelname training epochs \\
\modelname learning rate & $10^{-3}$ & $10^{-2}$ & $10^{-3}$ & $10^{-3}$ & $10^{-3}$ & learning rate for \modelname training \\
$\lambda$ & 0.5 & 0.05 & $10^{-3}$ & $10^{-3}$ & 0.5 & weight of negative log likelihood loss \\
$\beta$ & 0.1 & $10^{-3}$ & 0.5 & $10^{-3}$ & $5\times10^{-5}$ & weight of balance regularization \\
$M_{\mathrm{train}}$ & 3 & 6 & 3 & 3 & 8 & posterior samples during training \\
$M_{\mathrm{eval}}$ & 8 & 8 & 8 & 16 & 8 & posterior samples during evaluation \\

\addlinespace[0.5em]
\multicolumn{7}{l}{\textit{\textbf{(iii) Backbone \& optimization}}} \\
train epochs & 200 & 200 & 40 & 40 & 100 & number of backbone training epochs \\
learning rate & $5 \times 10^{-4}$ & $10^{-3}$ & $5 \times 10^{-4}$ & $10^{-3}$ & $10^{-3}$ & learning rate for backbone training \\
batch size & 32 & 128 & 128 & 128 & 32 & mini batch size \\
weight decay & $5 \times 10^{-4}$ & $5 \times 10^{-4}$ & $5 \times 10^{-4}$ & $5 \times 10^{-4}$ & $5 \times 10^{-4}$ & optimizer weight decay \\
\bottomrule
\end{tabular}}

{\raggedright \footnotesize \textit{Note:} M-ICU denotes MIMIC-IV (ICU delay) and M-Hosp denotes MIMIC-IV (in hospital).\par}
\end{table*}

\begin{table*}[t]
\centering
\caption{Ablation study of \modelname over 20 runs. 
Results with nominal coverage are shown in \textcolor{blue}{blue}, optimal values are \textbf{boldfaced}, and AvgSize exceeding half of the total number of categories is highlighted in \textcolor{mred}{red}. 
CovGap is reported in percent.}
\label{tab_ablation_results}
\setlength{\tabcolsep}{1.7pt} 
\renewcommand{\arraystretch}{1.1}
\resizebox{\linewidth}{!}{
\begin{tabular}{l ccc ccc ccc ccc}
\toprule
\multirow{2}{*}{\textbf{Method}} & \multicolumn{3}{c}{\textbf{SYN}} & \multicolumn{3}{c}{\textbf{Nursery}} & \multicolumn{3}{c}{\textbf{MIMIC}} & \multicolumn{3}{c}{\textbf{BACH}} \\
\cmidrule(lr){2-4} \cmidrule(lr){5-7} \cmidrule(lr){8-10} \cmidrule(lr){11-13}
 & Cov. & AvgSize  & CovGap  & Cov. & AvgSize  & CovGap  & Cov. & AvgSize  & CovGap & Cov. & AvgSize  & CovGap \\
\midrule
\textbf{Default \modelname} & \textbf{\textcolor{blue}{90.13$_{\pm 0.39}$}} & \textbf{2.225$_{\pm 0.035}$} & \textbf{4.2$_{\pm 0.7}$} & \textbf{\textcolor{blue}{91.65$_{\pm 0.70}$}} & 1.089$_{\pm 0.035}$ & \textbf{11.8$_{\pm 0.7}$} & \textbf{\textcolor{blue}{90.13$_{\pm 0.07}$}} & \textbf{1.107$_{\pm 0.001}$} & 0.8$_{\pm 0.0}$ & \textbf{\textcolor{blue}{91.90$_{\pm 2.70}$}} & \textbf{\textcolor{mred}{2.224$_{\pm 0.215}$}} & \textbf{6.2$_{\pm 1.1}$} \\
\midrule
(a) w/o Stochastic & 89.82$_{\pm 1.48}$ & 2.348$_{\pm 0.081}$ & 7.35$_{\pm 1.18}$ & \textcolor{blue}{91.38$_{\pm 1.15}$} & 1.123$_{\pm 0.051}$ & 13.42$_{\pm 1.76}$ & 89.94$_{\pm 0.96}$ & 1.053$_{\pm 0.013}$ & 2.36$_{\pm 0.48}$ & \textcolor{blue}{91.76$_{\pm 4.42}$} & \textcolor{mred}{2.447$_{\pm 0.308}$} & 8.84$_{\pm 2.35}$ \\
(b) Feature space & \textcolor{blue}{90.96$_{\pm 1.96}$} & 2.476$_{\pm 0.127}$ & 7.18$_{\pm 1.46}$ & \textcolor{blue}{90.92$_{\pm 1.43}$} & 1.248$_{\pm 0.079}$ & 14.16$_{\pm 2.05}$ & \textcolor{blue}{90.48$_{\pm 1.47}$} & 1.179$_{\pm 0.044}$ & 4.08$_{\pm 1.08}$ & 89.61$_{\pm 5.74}$ & \textcolor{mred}{3.094$_{\pm 0.446}$} & 12.41$_{\pm 3.16}$ \\
(c) w/o rank & 89.21$_{\pm 4.42}$ & 1.898$_{\pm 0.246}$ & 8.42$_{\pm 4.10}$ & 88.63$_{\pm 2.94}$ & 1.302$_{\pm 0.118}$ & 15.31$_{\pm 2.41}$ & \textcolor{blue}{90.03$_{\pm 5.41}$} & 1.334$_{\pm 0.146}$ & 5.49$_{\pm 3.71}$ & \textcolor{blue}{91.18$_{\pm 6.31}$} & \textcolor{mred}{3.491$_{\pm 0.541}$} & 9.26$_{\pm 3.02}$ \\
(d) w/o NLL & 88.36$_{\pm 1.17}$ & 2.074$_{\pm 0.108}$ & 5.39$_{\pm 1.08}$ & \textcolor{blue}{92.39$_{\pm 0.92}$} & 1.446$_{\pm 0.094}$ & 12.43$_{\pm 1.19}$ & \textcolor{blue}{90.02$_{\pm 2.96}$} & 1.351$_{\pm 0.078}$ & 2.96$_{\pm 0.77}$ & \textcolor{blue}{93.11$_{\pm 2.96}$} & \textcolor{mred}{2.846$_{\pm 0.317}$} & 7.44$_{\pm 1.47}$ \\
(e) w/o MI & \textcolor{blue}{92.28$_{\pm 3.41}$} & 2.438$_{\pm 0.241}$ & 6.53$_{\pm 2.72}$ & \textcolor{blue}{91.26$_{\pm 2.43}$} & 1.149$_{\pm 0.147}$ & 14.72$_{\pm 1.42}$ & 89.84$_{\pm 6.31}$ & 1.047$_{\pm 0.116}$ & 4.95$_{\pm 2.31}$ & \textcolor{blue}{90.91$_{\pm 7.82}$} & \textcolor{mred}{2.347$_{\pm 0.472}$} & 9.31$_{\pm 2.11}$ \\
\bottomrule
\end{tabular}
}
\end{table*}

\begin{table*}[t]
\centering
\caption{Performance comparison w.r.t. sample sparsity on the Synthetic Dataset (Target $1-\alpha=0.90$). \textbf{Takeaway:} \modelname remains consistently close to nominal coverage across all sample sizes while achieving the smallest subgroup coverage gap and comparatively small prediction sets. And the performance of each specific sensitive group is shown in the Fig.~\ref{fig_syn_group}.}
\label{tab_syn_sample}
\setlength{\tabcolsep}{1.2pt}
\renewcommand{\arraystretch}{1.1}
\resizebox{\linewidth}{!}{
\begin{tabular}{l ccc ccc ccc ccc}
\toprule
\multirow{2}{*}{\textbf{Method}} & \multicolumn{3}{c}{$\mathbf{N=500}$} & \multicolumn{3}{c}{$\mathbf{N=1000}$} & \multicolumn{3}{c}{$\mathbf{N=2000}$} & \multicolumn{3}{c}{$\mathbf{N=4000}$} \\
\cmidrule(lr){2-4} \cmidrule(lr){5-7} \cmidrule(lr){8-10} \cmidrule(lr){11-13}
 & Cov. $\uparrow$ & AvgSize $\downarrow$ & CovGap $\downarrow$ & Cov. $\uparrow$ & AvgSize $\downarrow$ & CovGap $\downarrow$ & Cov. $\uparrow$ & AvgSize $\downarrow$ & CovGap $\downarrow$ & Cov. $\uparrow$ & AvgSize $\downarrow$ & CovGap $\downarrow$ \\
\midrule
Marginal    & \textcolor{blue}{90.15$_{\pm 6.50}$} & 2.150$_{\pm 0.420}$ & 32.5$_{\pm 6.5}$ & \textcolor{blue}{90.20$_{\pm 4.20}$} & 1.950$_{\pm 0.400}$ & 21.5$_{\pm 6.2}$ & \textcolor{blue}{90.12$_{\pm 4.10}$} & 1.820$_{\pm 0.380}$ & 19.7$_{\pm 6.0}$ & 89.52$_{\pm 1.52}$ & 2.309$_{\pm 0.078}$ & 14.5$_{\pm 1.6}$ \\
Partial     & \textcolor{blue}{95.50$_{\pm 3.20}$} & \textcolor{mred}{4.520$_{\pm 0.300}$} & 9.5$_{\pm 2.8}$ & \textcolor{blue}{94.20$_{\pm 3.00}$} & 2.950$_{\pm 0.280}$ & 7.9$_{\pm 2.5}$ & \textcolor{blue}{93.50$_{\pm 2.80}$} & 2.820$_{\pm 0.250}$ & 7.5$_{\pm 2.2}$ & \textcolor{blue}{94.50$_{\pm 1.12}$} & 2.419$_{\pm 0.048}$ & 6.2$_{\pm 0.7}$ \\
Exhaustive  & \textcolor{blue}{100.00$_{\pm 0.00}$} & \textcolor{mred}{6.000$_{\pm 0.000}$} & 10.0$_{\pm 0.0}$ & \textcolor{blue}{98.50$_{\pm 2.20}$} & \textcolor{mred}{4.850$_{\pm 0.250}$} & 10.5$_{\pm 3.0}$ & \textcolor{blue}{98.20$_{\pm 2.10}$} & \textcolor{mred}{4.150$_{\pm 0.220}$} & 7.9$_{\pm 2.7}$ & \textcolor{blue}{95.85$_{\pm 0.91}$} & \textcolor{mred}{3.342$_{\pm 0.053}$} & 6.0$_{\pm 0.6}$ \\
\midrule
RLCP        & \textcolor{blue}{95.80$_{\pm 3.80}$} & 2.910$_{\pm 0.520}$ & 9.2$_{\pm 8.2}$ & \textcolor{blue}{95.10$_{\pm 3.20}$} & 2.850$_{\pm 0.480}$ & 8.5$_{\pm 6.8}$ & \textcolor{blue}{94.60$_{\pm 2.90}$} & 2.850$_{\pm 0.460}$ & 6.4$_{\pm 5.5}$ & \textcolor{blue}{93.10$_{\pm 2.45}$} & 2.540$_{\pm 0.305}$ & 4.5$_{\pm 1.8}$ \\
CluCP       & \textcolor{blue}{90.22$_{\pm 2.06}$} & \textcolor{mred}{3.112$_{\pm 0.332}$} & 10.7$_{\pm 0.9}$ & \textcolor{blue}{90.62$_{\pm 0.98}$} & 2.722$_{\pm 0.269}$ & 9.9$_{\pm 0.7}$ & \textcolor{blue}{91.11$_{\pm 1.83}$} & 2.756$_{\pm 0.250}$ & 8.9$_{\pm 1.1}$ & 89.80$_{\pm 1.78}$ & 2.426$_{\pm 0.050}$ & 4.9$_{\pm 0.8}$ \\
AFCP        & \textcolor{blue}{90.67$_{\pm 1.44}$} & \textcolor{mred}{3.162$_{\pm 0.292}$} & 6.4$_{\pm 0.8}$ & \textcolor{blue}{91.30$_{\pm 1.24}$} & 2.854$_{\pm 0.328}$ & 7.6$_{\pm 0.7}$ & \textcolor{blue}{91.62$_{\pm 1.71}$} & 2.848$_{\pm 0.227}$ & 7.3$_{\pm 0.9}$ & \textcolor{blue}{91.77$_{\pm 1.23}$} & 2.432$_{\pm 0.043}$ & 4.6$_{\pm 0.5}$ \\
FaReG       & \textcolor{blue}{92.18$_{\pm 1.26}$} & \textcolor{mred}{3.536$_{\pm 0.388}$} & 7.5$_{\pm 0.6}$ & \textcolor{blue}{92.41$_{\pm 1.13}$} & \textcolor{mred}{3.178$_{\pm 0.288}$} & 8.5$_{\pm 0.6}$ & \textcolor{blue}{92.49$_{\pm 1.57}$} & \textcolor{mred}{3.141$_{\pm 0.238}$} & 8.5$_{\pm 0.9}$ & \textcolor{blue}{92.61$_{\pm 0.98}$} & 2.579$_{\pm 0.029}$ & 4.7$_{\pm 0.5}$ \\
\midrule
\textbf{\modelname} & \textbf{\textcolor{blue}{91.20$_{\pm 1.20}$}} & \textbf{2.050$_{\pm 0.120}$} & \textbf{6.1$_{\pm 0.9}$} & \textbf{\textcolor{blue}{90.80$_{\pm 1.10}$}} & \textbf{1.820$_{\pm 0.110}$} & \textbf{5.1$_{\pm 0.8}$} & \textbf{\textcolor{blue}{90.50$_{\pm 1.00}$}} & \textbf{1.710$_{\pm 0.090}$} & \textbf{4.7$_{\pm 0.8}$} & \textbf{\textcolor{blue}{90.13$_{\pm 0.39}$}} & \textbf{2.225$_{\pm 0.035}$}& \textbf{4.2$_{\pm 0.7}$}\\
\bottomrule
\end{tabular}
}
\end{table*}

\begin{table}[t]
\centering
\caption{Dataset access and preprocessing sources.}
\label{tab_data_access}
\resizebox{\linewidth}{!}{
\begin{tabular}{|l|l|l|}
\hline
Dataset & Source & Access note \\
\hline
Synthetic & Procedurally generated; see Appendix~\ref{appendix_syn} 
& Generated from the described simulation protocol \\
\hline
Nursery & \texttt{https://archive.ics.uci.edu/dataset/76/nursery} 
& Public UCI dataset \\
\hline
MIMIC-IV & \texttt{https://physionet.org/content/mimiciv/3.1/} 
& Credentialed access via PhysioNet \\
\hline
BACH & \texttt{https://zenodo.org/records/3632035} 
& Public Zenodo record \\
\hline
\end{tabular}
}
\end{table}

\begin{table}[t]
\centering
\caption{Code availability and baseline implementations.}
\label{tab_code_access}
\resizebox{\linewidth}{!}{
\begin{tabular}{|l|l|}
\hline
Method & Code source \\
\hline
\modelname & \texttt{https://anonymous.4open.science/r/sgcp/} \\
\hline
Partial Equalized & \texttt{https://github.com/yromano/cqr} \\
\hline
RLCP & \texttt{https://github.com/rohanhore/RLCP} \\
\hline
CluCP & \texttt{https://github.com/tiffanyding/class-conditional-conformal} \\
\hline
AFCP & \texttt{https://github.com/FionaZ3696/Adaptively-Fair-Conformal-Prediction} \\
\hline
FaReG & \texttt{https://github.com/Xusr1123/FaReG} \\
\hline
\end{tabular}
}
\end{table}

\paragraph{BACH.}
The Breast Cancer Histology (BACH) dataset~\citep{ARESTA2019122} is a histopathology benchmark for evaluating subgroup reliability under visual heterogeneity. 
We use Part~A, which contains 400 H\&E-stained microscopy images evenly distributed across four classes: normal, benign, in situ carcinoma, and invasive carcinoma. 
Images were acquired at $20\times$ magnification with resolution $2048\times1536$, annotated by two pathologists, and collected from 39 patients, though anonymization prevents complete patient-level tracing. 
The task is four-class classification with
\[
Y\in\{\mathrm{Normal},\mathrm{Benign},\mathrm{In\ Situ\ Carcinoma},\mathrm{Invasive\ Carcinoma}\}.
\]

Since demographic sensitive attributes are unavailable, we use the class label as a proxy grouping variable and set $A:=Y$. 
Subgroup reliability is therefore evaluated across disease categories, which naturally differ in visual complexity, intra-class variability, and class-boundary ambiguity. 
For preprocessing, images are resized to $224\times224$, augmented during training with random horizontal and vertical flips, and normalized using ImageNet statistics. 
We train a visual backbone $\Phi$ and use the learned representations $X_i=\Phi(I_i)\in\mathbb R^d$ for downstream prediction and evaluation.

In this benchmark, heterogeneity is primarily visual rather than demographic. 
Differences in tissue morphology, staining, lesion structure, and acquisition conditions can induce distinct class-conditional score laws, $F_c(s):=\mathbb P(S\le s\mid Y=c)$.
Thus, BACH complements the tabular benchmarks by testing whether \modelname can maintain reliable conditional coverage when subgroup structure reflects semantic visual heterogeneity rather than explicit sensitive attributes.

\subsection{Implementation Details}\label{appendix_para}

\paragraph{Computing resources.}
All experiments are conducted on a workstation equipped with an Intel i5-14600KF CPU, 32 GB of RAM, and a single NVIDIA GeForce RTX 4090 D GPU. 
All models are implemented in PyTorch, and each experimental setting is trained with multiple random seeds. 
The hyperparameters for each dataset are reported in Table~\ref{tab_hparams}. 
Data access and code availability are summarized in Tables~\ref{tab_data_access} and~\ref{tab_code_access}, respectively.

\paragraph{Open access to data and code.}
All external datasets used in our experiments are publicly documented and accessible online, subject to their original access requirements. 
The synthetic dataset is generated procedurally following Appendix~\ref{appendix_syn}. 
The code for \modelname is provided anonymously in the supplementary material, and baseline implementations are based on the publicly available repositories listed below.

\subsubsection{Baseline implementation details}
We compare \modelname\ with six baselines: Partial Equalized, Exhaustive Equalized, CluCP, AFCP, FaReG, and RLCP. 
Unless otherwise stated, all methods use the same backbone classifier and softmax-based nonconformity score as \modelname. 
Hyperparameters are adopted from the original papers or released implementations whenever available.

\textbf{Partial and Exhaustive Equalized.}
We extend the equalized-coverage framework of~\cite{Romano2020With} to conformal prediction by applying group-conditional calibration over sensitive groups. 
Partial Equalized calibrates separately for each value of one sensitive attribute, whereas Exhaustive Equalized calibrates over all intersectional groups induced by the full combination of sensitive attributes. 
In both cases, the classifier is trained on all samples, and only the calibration step is group-conditional.

\textbf{CluCP.}
We implement CluCP following its original clustered conformal pipeline. 
The clustering fraction $\gamma$ and the number of clusters $M$ are selected using the heuristic proposed in the paper. 
Class embeddings are clustered by weighted $k$-means with the default \texttt{KMeans} implementation. 
Although CluCP targets class-wise rather than subgroup-fair calibration, it provides a useful non-global calibration baseline.

\textbf{AFCP.}
We implement AFCP using its original adaptive attribute-selection procedure, directly applying it to the sensitive attributes available in each dataset.

\textbf{FaReG.}
We follow the original FaReG representation-learning pipeline: an encoder-decoder first learns latent representations, representation-based low-coverage groups are then identified, and prediction sets are adjusted accordingly. 
We use the original three-layer MLP encoder and decoder. 
For Synthetic and Nursery, we adopt the reported hyperparameters: hidden dimensions $[64,32]$, batch size $500$, learning rates $10^{-3}$ and $10^{-2}$, respectively, with $(\beta,\delta,T)=(2.0,0.3,20)$ for Synthetic and $(0.1,0.1,100)$ for Nursery. 
For the remaining datasets, we keep the architecture and optimization settings as close as possible to the released implementation.

\textbf{RLCP.}
We implement RLCP following its weighted localization procedure. 
Calibration samples are weighted by their similarity to each test point, and prediction sets are constructed using the corresponding weighted conformal quantile. 
For all datasets, we use the Gaussian kernel and the bandwidth-selection rule from the original method, without subgroup-specific tuning.

\section{Additional Experiments}\label{appendix_exp}

\begin{figure}[t]
    \centering
    \includegraphics[width=0.8\linewidth]{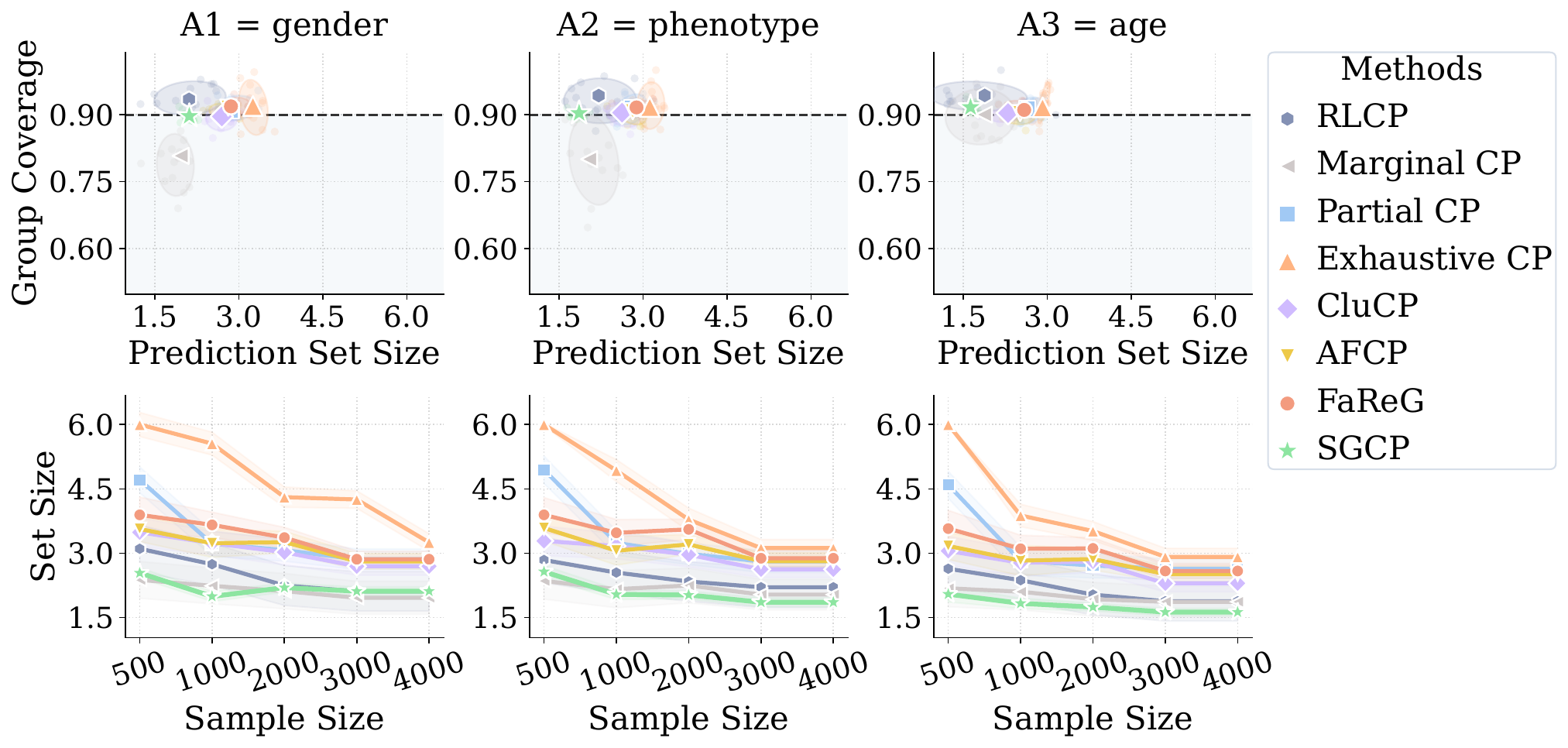}
    \caption{Group coverage on SYN dataset. \modelname remains close to the target coverage across classes while using smaller prediction sets than more conservative baselines.}
    \label{fig_syn_group}
\end{figure}

\begin{figure}[t]
    \centering
    \includegraphics[width=0.8\linewidth]{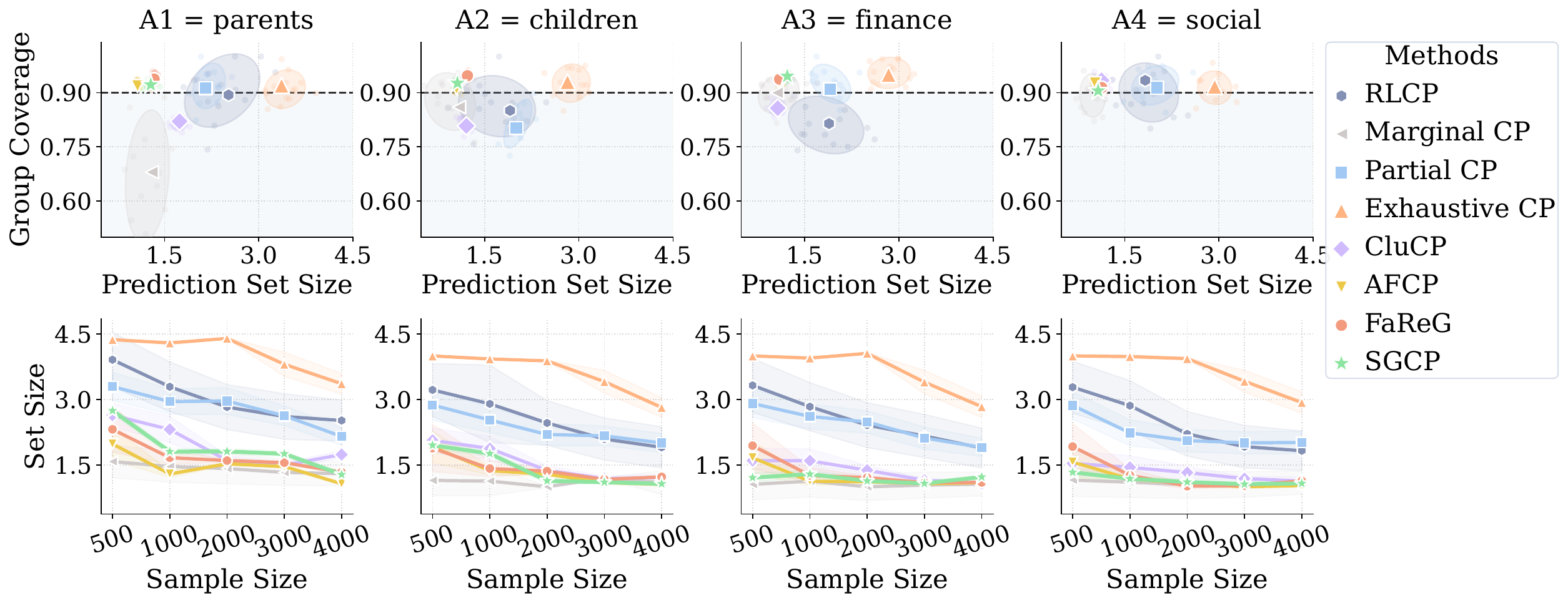}
    \caption{Group coverage on Nursery. Top: subgroup coverage \emph{vs.} average prediction set size at a sample size of 4000, with the dashed line indicating nominal coverage of $0.9$. Bottom: prediction set size varies with subgroup sample size. \modelname stays close to the nominal target while remaining among the most efficient methods across subgroup definitions.}
    \vspace{-2mm}
    \label{fig_nursery_group}
\end{figure}

\begin{table*}[t]
\centering
\caption{Performance comparison w.r.t. sample sparsity on the Nursery Dataset (Target $1-\alpha=0.90$). \textbf{Takeaway:} \modelname remains near nominal coverage even in the sparse regime and consistently achieves the smallest subgroup coverage gap, while competing methods improve subgroup reliability only with larger prediction sets or at the cost of under- or over-coverage. And the performance of each specific sensitive group is shown in the Fig.~\ref{fig_nursery_group}.}
\label{tab_nursery_sample}
\setlength{\tabcolsep}{1.5pt}
\renewcommand{\arraystretch}{1.1}
\resizebox{\linewidth}{!}{
\begin{tabular}{l ccc ccc ccc ccc}
\toprule
\multirow{2}{*}{\textbf{Method}} & \multicolumn{3}{c}{$\mathbf{N=500}$} & \multicolumn{3}{c}{$\mathbf{N=1000}$} & \multicolumn{3}{c}{$\mathbf{N=2000}$} & \multicolumn{3}{c}{$\mathbf{N=4000}$} \\
\cmidrule(lr){2-4} \cmidrule(lr){5-7} \cmidrule(lr){8-10} \cmidrule(lr){11-13}
 & Cov. $\uparrow$ & AvgSize $\downarrow$ & CovGap $\downarrow$ & Cov. $\uparrow$ & AvgSize $\downarrow$ & CovGap $\downarrow$ & Cov. $\uparrow$ & AvgSize $\downarrow$ & CovGap $\downarrow$ & Cov. $\uparrow$ & AvgSize $\downarrow$ & CovGap $\downarrow$ \\
\midrule
Marginal    & 48.25$_{\pm 8.50}$ & 1.120$_{\pm 0.350}$ & 29.5$_{\pm 6.5}$ & 57.59$_{\pm 8.20}$ & 1.080$_{\pm 0.330}$ & 29.1$_{\pm 5.2}$ & 59.72$_{\pm 7.80}$ & 1.050$_{\pm 0.310}$ & 26.2$_{\pm 5.8}$ & 84.15$_{\pm 5.20}$ & \textcolor{mred}{2.170$_{\pm 0.260}$} & 20.4$_{\pm 5.2}$ \\
Partial     & 84.61$_{\pm 4.80}$ & \textcolor{mred}{2.850$_{\pm 0.310}$} & 17.7$_{\pm 3.8}$ & 86.07$_{\pm 4.50}$ & \textcolor{mred}{2.620$_{\pm 0.280}$} & 16.8$_{\pm 3.5}$ & 87.86$_{\pm 4.20}$ & \textcolor{mred}{2.450$_{\pm 0.250}$} & 15.2$_{\pm 3.2}$ & 88.97$_{\pm 3.80}$ & 1.850$_{\pm 0.200}$ & 12.1$_{\pm 2.5}$ \\
Exhaustive  & \textcolor{blue}{100.00$_{\pm 0.00}$} & \textcolor{mred}{4.000$_{\pm 0.000}$} & 10.0$_{\pm 0.0}$ & \textcolor{blue}{98.00$_{\pm 0.10}$} & \textcolor{mred}{3.980$_{\pm 0.001}$} & 18.8$_{\pm 0.2}$ & \textcolor{blue}{98.00$_{\pm 0.12}$} & \textcolor{mred}{3.979$_{\pm 0.001}$} & 17.5$_{\pm 0.8}$ & \textcolor{blue}{96.80$_{\pm 3.10}$} & \textcolor{mred}{2.850$_{\pm 0.220}$} & 15.5$_{\pm 3.2}$ \\
\midrule
RLCP        & 82.45$_{\pm 14.20}$ & \textcolor{mred}{2.250$_{\pm 0.580}$} & 18.5$_{\pm 9.5}$ & 83.36$_{\pm 13.80}$ & 1.720$_{\pm 0.550}$ & 17.8$_{\pm 6.2}$ & 83.31$_{\pm 11.50}$ & 1.480$_{\pm 0.510}$ & 16.5$_{\pm 5.8}$ & 88.67$_{\pm 6.20}$ & 1.220$_{\pm 0.410}$ & 13.5$_{\pm 3.8}$ \\
CluCP       & 86.26$_{\pm 5.61}$ & 1.506$_{\pm 0.265}$ & 17.3$_{\pm 4.5}$ & 83.80$_{\pm 3.59}$ & 1.516$_{\pm 0.265}$ & 17.1$_{\pm 2.9}$ & 84.42$_{\pm 2.47}$ & 1.261$_{\pm 0.096}$ & 16.9$_{\pm 1.8}$ & 85.08$_{\pm 1.60}$ & 1.121$_{\pm 0.064}$ & 12.8$_{\pm 1.3}$ \\
AFCP        & \textcolor{blue}{92.78$_{\pm 4.81}$} & 1.551$_{\pm 0.345}$ & 18.7$_{\pm 3.9}$ & \textcolor{blue}{94.85$_{\pm 3.69}$} & \textbf{1.163$_{\pm 0.074}$} & 16.4$_{\pm 2.8}$ & \textcolor{blue}{93.12$_{\pm 1.33}$} & \textbf{1.092$_{\pm 0.031}$} & 15.6$_{\pm 1.1}$ & \textcolor{blue}{92.85$_{\pm 0.84}$} & \textbf{1.085$_{\pm 0.024}$} & 14.1$_{\pm 0.6}$ \\
FaReG       & \textcolor{blue}{93.77$_{\pm 3.66}$} & 1.879$_{\pm 0.515}$ & 17.4$_{\pm 2.9}$ & \textcolor{blue}{92.47$_{\pm 3.04}$} & 1.346$_{\pm 0.146}$ & 16.2$_{\pm 2.3}$ & \textcolor{blue}{94.37$_{\pm 1.44}$} & 1.257$_{\pm 0.027}$ & 15.4$_{\pm 1.1}$ & \textcolor{blue}{93.21$_{\pm 0.62}$} & 1.230$_{\pm 0.012}$ & 13.3$_{\pm 0.6}$ \\
\midrule
\textbf{\modelname} & \textbf{\textcolor{blue}{92.72$_{\pm 1.50}$}} & \textbf{1.320$_{\pm 0.080}$} & \textbf{15.1$_{\pm 1.5}$} & \textbf{\textcolor{blue}{92.10$_{\pm 1.20}$}} & 1.210$_{\pm 0.060}$ & \textbf{14.6$_{\pm 1.2}$} & \textbf{\textcolor{blue}{91.97$_{\pm 1.00}$}} & 1.120$_{\pm 0.050}$ & \textbf{12.3$_{\pm 1.0}$} & \textbf{\textcolor{blue}{91.65$_{\pm 0.70}$}} & 1.089$_{\pm 0.035}$ & \textbf{11.8$_{\pm 0.7}$} \\
\bottomrule
\end{tabular}
}
\end{table*}

\begin{table*}[t]
\centering
\caption{Performance comparison w.r.t. sample sparsity on MIMIC-IV (target $1-\alpha=0.90$). Each method is run 20 times. \textbf{Takeaway:} On the MIMIC-IV dataset, \modelname consistently achieves near-nominal coverage while maintaining the lowest subpopulation disparity (CovGap) in sparse settings ($N=2000, 5000$). Even with the full dataset, \modelname provides a highly efficient prediction set with excellent fairness metrics compared to conventional baselines.}
\label{tab_mimic_sample}
\setlength{\tabcolsep}{3.7pt}
\renewcommand{\arraystretch}{1.1}
\resizebox{\linewidth}{!}{
\begin{tabular}{l ccc ccc ccc}
\toprule
\multirow{2}{*}{\textbf{Method}} & \multicolumn{3}{c}{$\mathbf{N=2000}$} & \multicolumn{3}{c}{$\mathbf{N=5000}$} & \multicolumn{3}{c}{\textbf{Full (All Samples)}} \\
\cmidrule(lr){2-4} \cmidrule(lr){5-7} \cmidrule(lr){8-10}
 & Cov. $\uparrow$ & AvgSize $\downarrow$ & CovGap $\downarrow$ & Cov. $\uparrow$ & AvgSize $\downarrow$ & CovGap $\downarrow$ & Cov. $\uparrow$ & AvgSize $\downarrow$ & CovGap $\downarrow$ \\
\midrule
Marginal    & \textcolor{blue}{90.06$_{\pm 10.14}$} & 1.074$_{\pm 0.211}$ & 34.4$_{\pm 10.9}$ & \textcolor{blue}{90.79$_{\pm 10.62}$} & \textbf{1.026$_{\pm 0.087}$} & 29.6$_{\pm 9.3}$ & \textcolor{blue}{90.79$_{\pm 10.62}$} & 1.226$_{\pm 0.087}$ & 27.1$_{\pm 7.1}$ \\
Partial     & \textcolor{blue}{95.73$_{\pm 3.62}$} & \textcolor{mred}{1.564$_{\pm 0.243}$} & 11.0$_{\pm 5.6}$ & \textcolor{blue}{94.50$_{\pm 4.54}$} & 1.335$_{\pm 0.207}$ & 9.2$_{\pm 4.0}$ & \textcolor{blue}{94.50$_{\pm 4.54}$} & 1.258$_{\pm 0.140}$ & 8.0$_{\pm 3.0}$ \\
Exhaustive  & \textcolor{blue}{100.00$_{\pm 0.00}$} & \textcolor{mred}{3.000$_{\pm 0.000}$} & 10.0$_{\pm 0.0}$ & \textcolor{blue}{98.86$_{\pm 5.21}$} & \textcolor{mred}{2.989$_{\pm 0.004}$} & 21.4$_{\pm 5.3}$ & \textcolor{blue}{96.45$_{\pm 4.22}$} & \textcolor{mred}{2.800$_{\pm 0.142}$} & 19.7$_{\pm 3.2}$ \\
\midrule
RLCP        & \textcolor{blue}{95.10$_{\pm 2.40}$} & \textcolor{mred}{1.650$_{\pm 0.120}$} & 4.2$_{\pm 2.6}$ & \textcolor{blue}{93.20$_{\pm 2.10}$} & 1.480$_{\pm 0.150}$ & 3.5$_{\pm 1.4}$ & \textcolor{blue}{92.50$_{\pm 2.80}$} & 1.350$_{\pm 0.100}$ & 1.7$_{\pm 0.8}$ \\
CluCP       & \textcolor{blue}{95.11$_{\pm 1.92}$} & \textcolor{mred}{1.654$_{\pm 0.192}$} & 4.5$_{\pm 1.5}$ & \textcolor{blue}{93.19$_{\pm 1.28}$} & 1.422$_{\pm 0.059}$ & 3.5$_{\pm 1.0}$ & \textcolor{blue}{92.22$_{\pm 1.28}$} & 1.302$_{\pm 0.059}$ & 3.5$_{\pm 0.8}$ \\
AFCP        & \textcolor{blue}{91.81$_{\pm 2.51}$} & 1.400$_{\pm 0.075}$ & 3.9$_{\pm 1.0}$ & \textcolor{blue}{91.40$_{\pm 1.12}$} & 1.114$_{\pm 0.045}$ & 3.3$_{\pm 0.8}$ & - & - & - \\
FaReG       & \textcolor{blue}{95.78$_{\pm 2.09}$} & \textcolor{mred}{1.601$_{\pm 0.078}$} & 4.1$_{\pm 1.2}$ & \textcolor{blue}{92.58$_{\pm 1.05}$} & 1.357$_{\pm 0.050}$ & 3.0$_{\pm 1.1}$ & \textcolor{blue}{90.23$_{\pm 0.09}$} & 1.174$_{\pm 0.001}$ & \textbf{0.5$_{\pm 0.0}$} \\
\midrule
\textbf{\modelname} & \textbf{\textcolor{blue}{90.12$_{\pm 1.91}$}} & \textbf{1.060$_{\pm 0.068}$} & \textbf{3.3$_{\pm 0.7}$} & \textbf{\textcolor{blue}{90.48$_{\pm 1.36}$}} & 1.116$_{\pm 0.050}$ & \textbf{2.6$_{\pm 0.5}$} & \textbf{\textcolor{blue}{90.13$_{\pm 0.07}$}} & \textbf{1.107$_{\pm 0.001}$} & 0.8$_{\pm 0.0}$ \\
\bottomrule
\end{tabular}
}
\end{table*}

\subsection{Ablation Study}\label{subsec:ablations}
Table~\ref{tab_ablation_results} studies the contribution of each component in \modelname.
The default model achieves nominal marginal coverage on all datasets and obtains the best or tied-best CovGap on three out of four datasets, while maintaining the smallest or near-smallest prediction sets.
This suggests that the full objective provides the most favorable balance between validity, subgroup reliability, and efficiency.

Removing posterior stochasticity consistently degrades subgroup reliability.
The deterministic variant increases CovGap on all datasets, indicating that posterior averaging is useful for stabilizing the local score law rather than committing each sample to a single calibration state.
Replacing the learned calibration state with feature-space grouping also leads to larger CovGap and larger prediction sets, especially on MIMIC and BACH.
This supports our motivation that similarity in raw feature space is not necessarily the right notion of locality for conformal calibration; what matters is similarity in score behavior.

The loss ablations further clarify the role of each training signal.
Without the rank-uniformization loss, coverage becomes less stable and CovGap increases substantially, showing that aligning transformed scores to a common percentile scale is central to subgroup reliability.
Removing the density-fitting term also worsens the tradeoff between coverage and efficiency, often producing larger prediction sets, which suggests that likelihood-based fitting helps the learned local law remain faithful to the observed score distribution.
Finally, removing the mutual-information term leads to less reliable subgroup calibration and higher variance across runs, consistent with component underuse or less informative stochastic grouping.
Overall, the ablation results support the design of \modelname: subgroup reliability improves when local score laws are learned through both distributional fitting and rank-based normalization, and when calibration evidence is shared through informative stochastic components.

\begin{figure}[t]
    \centering
    \includegraphics[width=0.8\linewidth]{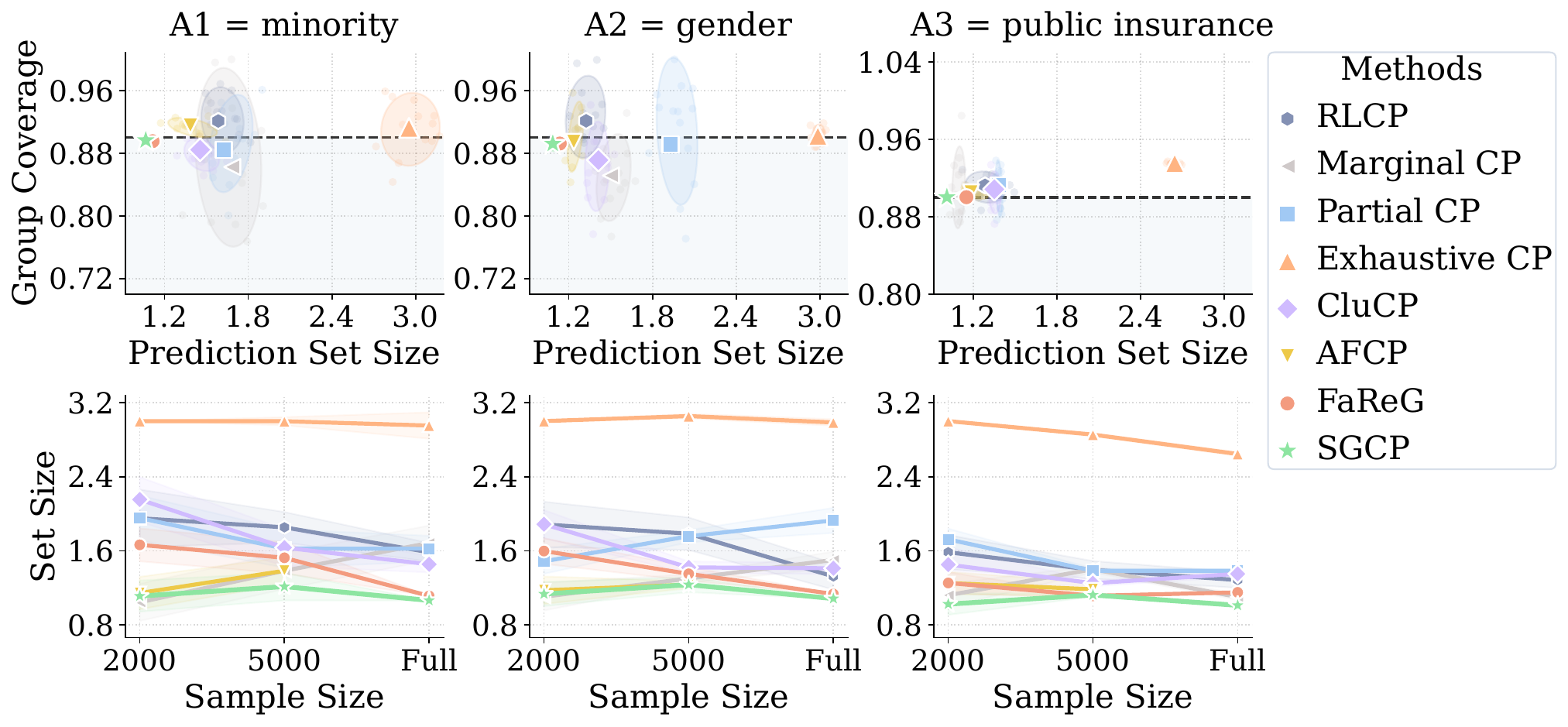}
    \caption{Group coverage on MIMIC-IVdataset. \modelname remains close to the target coverage across classes while using smaller prediction sets than more conservative baselines.}
    \vspace{-2mm}
    \label{fig_mimic_group}
\end{figure}

\begin{table*}[t]
\centering
\caption{Performance comparison w.r.t. sample sparsity on BACH (target $1-\alpha=0.90$). Each method is run 20 times. \textbf{Takeaway:} Under extreme sparsity, most methods become conservative, but with the increase in sample size, \modelname recovers the fastest, with near-nominal coverage, a smaller prediction set, and lower subpopulation disparity, outperforming other baseline methods. The performance of each individual subgroup is further shown in Fig.~\ref{fig_bach_group}.}
\label{tab_bach_sample}
\setlength{\tabcolsep}{1.2pt}
\renewcommand{\arraystretch}{1.1}
\resizebox{\linewidth}{!}{
\begin{tabular}{l ccc ccc ccc ccc}
\toprule
\multirow{2}{*}{\textbf{Method}} & \multicolumn{3}{c}{$\mathbf{N=50}$} & \multicolumn{3}{c}{$\mathbf{N=150}$} & \multicolumn{3}{c}{$\mathbf{N=200}$} & \multicolumn{3}{c}{$\mathbf{N=300}$} \\
\cmidrule(lr){2-4} \cmidrule(lr){5-7} \cmidrule(lr){8-10} \cmidrule(lr){11-13}
 & Cov. $\uparrow$ & AvgSize $\downarrow$ & CovGap $\downarrow$ & Cov. $\uparrow$ & AvgSize $\downarrow$ & CovGap $\downarrow$ & Cov. $\uparrow$ & AvgSize $\downarrow$ & CovGap $\downarrow$ & Cov. $\uparrow$ & AvgSize $\downarrow$ & CovGap $\downarrow$ \\
\midrule
Marginal    & 89.34$_{\pm 3.35}$ & \textcolor{mred}{3.700$_{\pm 0.474}$} & 19.0$_{\pm 2.1}$ & \textcolor{blue}{90.50$_{\pm 3.86}$} & \textcolor{mred}{3.604$_{\pm 0.494}$} & 17.6$_{\pm 2.0}$ & \textcolor{blue}{90.69$_{\pm 15.75}$} & \textcolor{mred}{3.545$_{\pm 0.445}$} & 15.9$_{\pm 2.9}$ & 89.20$_{\pm 9.64}$ & \textcolor{mred}{2.894$_{\pm 0.267}$} & 15.7$_{\pm 1.7}$ \\
Partial     & \textcolor{blue}{98.33$_{\pm 2.66}$} & \textcolor{mred}{3.735$_{\pm 0.316}$} & 7.8$_{\pm 1.0}$ & \textcolor{blue}{97.45$_{\pm 3.25}$} & \textcolor{mred}{3.711$_{\pm 0.290}$} & \textbf{6.2$_{\pm 0.6}$} & \textcolor{blue}{97.17$_{\pm 11.56}$} & \textcolor{mred}{3.460$_{\pm 0.258}$} & 7.5$_{\pm 0.7}$ & \textcolor{blue}{96.57$_{\pm 8.08}$} & \textcolor{mred}{3.385$_{\pm 0.299}$} & 7.6$_{\pm 1.6}$ \\
Exhaustive  & \textcolor{blue}{100.00$_{\pm 0.00}$} & \textcolor{mred}{4.000$_{\pm 0.00}$} & 10.0$_{\pm 0.0}$ & \textcolor{blue}{100.00$_{\pm 0.0}$} & \textcolor{mred}{4.000$_{\pm 0.0}$} & 10.0$_{\pm 0.0}$ & \textcolor{blue}{98.73$_{\pm 0.63}$} & \textcolor{mred}{3.837$_{\pm 0.233}$} & 11.0$_{\pm 0.1}$ & \textcolor{blue}{97.41$_{\pm 6.89}$} & \textcolor{mred}{3.867$_{\pm 0.118}$} & 11.6$_{\pm 1.5}$ \\
\midrule
RLCP        & \textcolor{blue}{91.74$_{\pm 3.06}$} & \textbf{\textcolor{mred}{2.855$_{\pm 0.440}$}} & 8.2$_{\pm 3.6}$ & \textcolor{blue}{92.25$_{\pm 3.80}$} & \textbf{\textcolor{mred}{2.763$_{\pm 0.422}$}} & 7.9$_{\pm 3.8}$ & \textcolor{blue}{91.99$_{\pm 10.88}$} & \textcolor{mred}{2.611$_{\pm 0.411}$} & 6.4$_{\pm 3.8}$ & \textcolor{blue}{91.88$_{\pm 10.56}$} & \textcolor{mred}{2.311$_{\pm 0.319}$} & 8.1$_{\pm 2.5}$ \\
CluCP       & \textcolor{blue}{97.80$_{\pm 3.14}$} & \textcolor{mred}{3.936$_{\pm 0.199}$} & 8.7$_{\pm 1.5}$ & \textcolor{blue}{96.60$_{\pm 3.29}$} & \textcolor{mred}{3.882$_{\pm 0.180}$} & 8.1$_{\pm 1.9}$ & \textcolor{blue}{98.34$_{\pm 0.50}$} & \textcolor{mred}{3.985$_{\pm 0.110}$} & 10.0$_{\pm 2.6}$ & \textcolor{blue}{95.91$_{\pm 3.59}$} & \textcolor{mred}{3.257$_{\pm 0.220}$} & 10.7$_{\pm 2.6}$ \\
AFCP        & \textcolor{blue}{100.00$_{\pm 0.50}$} & \textcolor{mred}{4.000$_{\pm 0.120}$} & 10.0$_{\pm 1.2}$ & \textcolor{blue}{96.50$_{\pm 2.64}$} & \textcolor{mred}{3.817$_{\pm 0.116}$} & 8.3$_{\pm 1.3}$ & \textcolor{blue}{95.80$_{\pm 2.39}$} & \textcolor{mred}{3.447$_{\pm 0.297}$} & 7.5$_{\pm 1.2}$ & \textcolor{blue}{95.00$_{\pm 2.60}$} & \textcolor{mred}{2.830$_{\pm 0.237}$} & 6.3$_{\pm 1.7}$ \\
FaReG       & \textcolor{blue}{99.00$_{\pm 2.00}$} & \textcolor{mred}{3.946$_{\pm 0.153}$} & 9.3$_{\pm 1.3}$ & \textcolor{blue}{98.40$_{\pm 1.78}$} & \textcolor{mred}{3.914$_{\pm 0.212}$} & 8.4$_{\pm 1.8}$ & \textcolor{blue}{96.80$_{\pm 1.93}$} & \textcolor{mred}{3.769$_{\pm 0.250}$} & 7.7$_{\pm 1.5}$ & \textcolor{blue}{93.60$_{\pm 3.40}$} & \textcolor{mred}{2.959$_{\pm 0.239}$} & 7.4$_{\pm 1.5}$ \\
\midrule
\textbf{\modelname} & \textcolor{blue}{96.44$_{\pm 2.61}$} & \textcolor{mred}{3.842$_{\pm 0.136}$} & \textbf{6.8$_{\pm 2.2}$} & \textcolor{blue}{92.63$_{\pm 4.04}$} & \textcolor{mred}{3.344$_{\pm 0.240}$} & \textbf{5.0$_{\pm 1.58}$} & \textcolor{blue}{92.22$_{\pm 2.39}$} & \textbf{\textcolor{mred}{2.370$_{\pm 0.197}$}} & \textbf{4.6$_{\pm 2.1}$} & \textbf{\textcolor{blue}{91.90$_{\pm 2.70}$}} & \textbf{\textcolor{mred}{2.224$_{\pm 0.215}$}} & \textbf{6.2$_{\pm 1.1}$} \\
\bottomrule
\end{tabular}
}
\end{table*}
\begin{figure}[ht]
    \centering
    \includegraphics[width=1\linewidth]{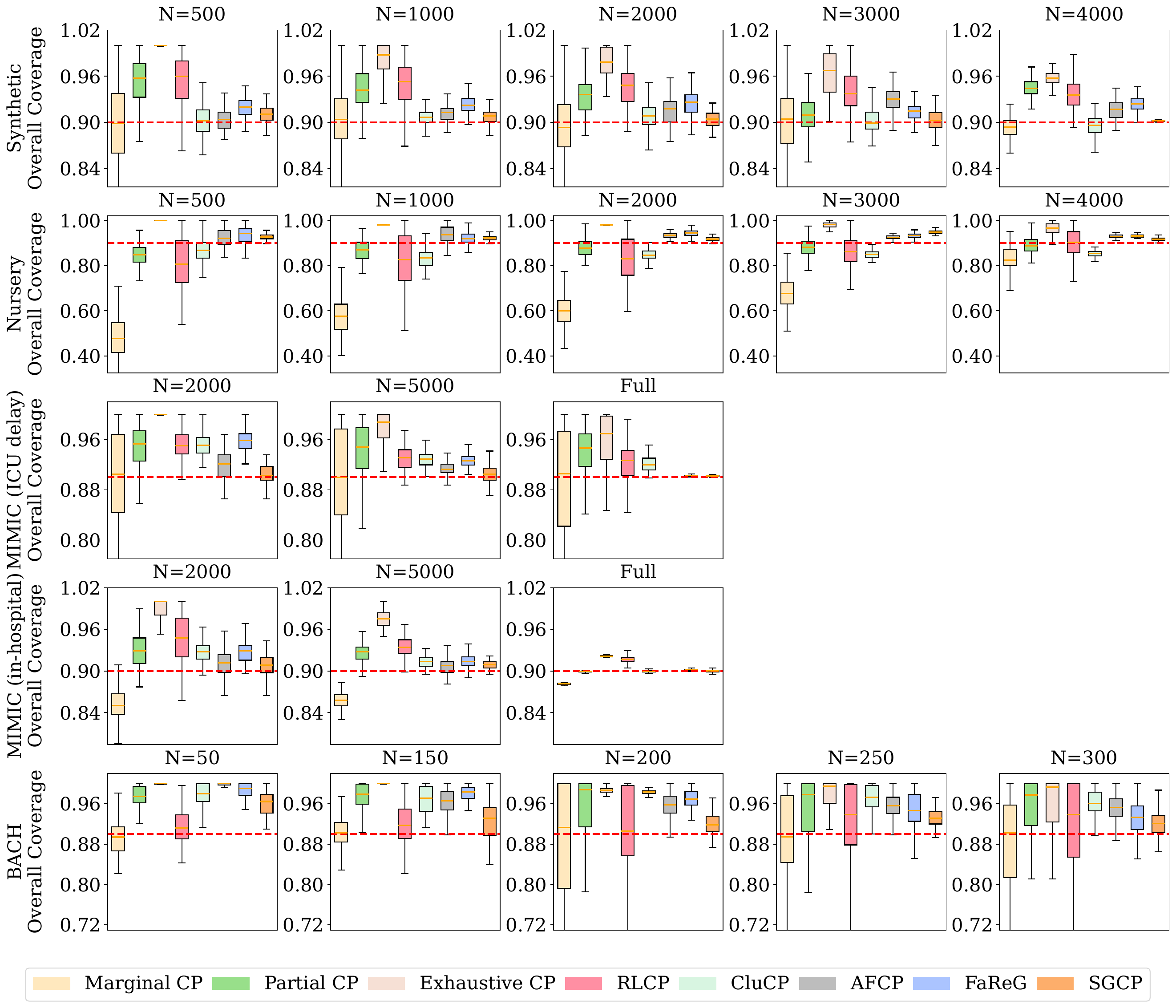}
    \vspace{-2mm}
    \caption{Distribution of overall coverage over repeated runs under varying sample sparsity. The dashed line marks the nominal target coverage of $0.9$. \textbf{Takeaway:} \modelname remains consistently close to the target across datasets and sample sizes, whereas global calibration may under-cover in sparse regimes and stronger equalization baselines often become overly conservative.}
    \vspace{-2mm}
    \label{fig_coverage_hist}
\end{figure}

\begin{figure}[ht]
    \centering
    \includegraphics[width=1\linewidth]{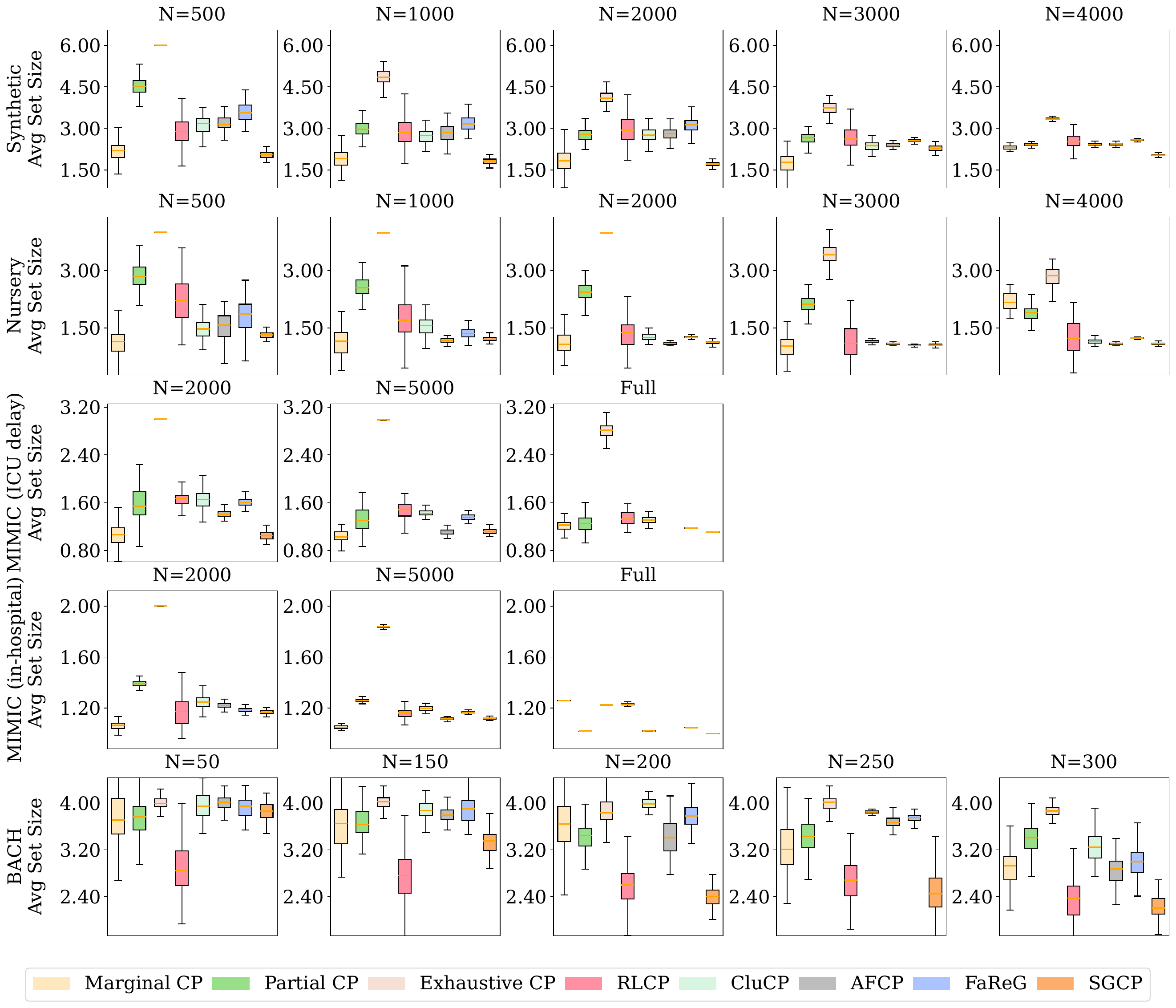}
    \caption{Distribution of average prediction set size over repeated runs under varying sample sparsity. Smaller values indicate better efficiency. \textbf{Takeaway:} \modelname remains among the most efficient methods across datasets and sample sizes, often matching or improving upon the smallest prediction sets while avoiding the severe conservatism of explicit equalization and localized baselines.}
    \label{fig_size_hist}
\end{figure}

\subsection{Additional Biomedical Benchmarks}
\label{app:additional_biomedical}

\begin{table}[t]
\centering
\caption{Performance comparison on the additional biomedical dataset HAM10000 at 90\% nominal coverage. \textbf{Takeaway:} \modelname achieves the lowest intersectional CovGap while maintaining near-nominal coverage and compact prediction sets.}
\label{tab:ham10000_additional}
\begin{tabular}{lccc}
\toprule
Method & Coverage (\%) & AvgSize & Intersection CovGap \\
\midrule
Marginal CP & $90.25 \pm 0.88$ & $1.35 \pm 0.05$ & $7.30 \pm 0.62$ \\
Partial & $90.24 \pm 0.90$ & $1.34 \pm 0.05$ & $7.40 \pm 0.54$ \\
AFCP & $90.44 \pm 0.86$ & $1.64 \pm 0.10$ & $6.20 \pm 0.49$ \\
FaReG & $93.02 \pm 1.09$ & $1.93 \pm 0.15$ & $8.00 \pm 1.09$ \\
\modelname & $90.30 \pm 0.88$ & $1.36 \pm 0.05$ & $\mathbf{5.99 \pm 0.84}$ \\
\bottomrule
\end{tabular}
\end{table}
\begin{table}[t]
\centering
\caption{Performance comparison on the additional biomedical dataset MIMIC-CXR. \textbf{Takeaway:} \modelname matches Marginal CP in average set size and achieves competitive intersectional reliability.}
\label{tab:mimiccxr_additional}
\begin{tabular}{lccc}
\toprule
Method & Coverage (\%) & AvgSize & Intersection CovGap \\
\midrule
Marginal CP & $90.27 \pm 0.70$ & $1.39 \pm 0.34$ & $1.24 \pm 0.34$ \\
Partial & $90.49 \pm 0.63$ & $1.41 \pm 0.02$ & $1.35 \pm 0.33$ \\
AFCP & $90.31 \pm 0.86$ & $1.49 \pm 0.02$ & $\mathbf{1.14 \pm 0.32}$ \\
FaReG & $93.78 \pm 0.59$ & $1.62 \pm 0.03$ & $3.78 \pm 0.59$ \\
\modelname & $90.25 \pm 0.72$ & $1.39 \pm 0.02$ & $1.21 \pm 0.33$ \\
\bottomrule
\end{tabular}
\end{table}

To further assess the generalizability of \modelname across biomedical imaging tasks, we conduct additional experiments on two datasets, HAM10000 and MIMIC-CXR. These experiments extend our evaluation to dermatology and chest radiography, respectively.

\textbf{Datasets.}
HAM10000~\cite{Tschandl2018} is a public dermoscopic image dataset containing 10,015 images from seven diagnostic categories. The associated metadata include sex, age, and anatomical lesion site, enabling subgroup evaluation based on individual attributes and their intersections. MIMIC-CXR~\cite{PhysioNet-mimic-cxr-2.0.0, Johnson2019} is a large-scale clinical chest radiograph dataset with associated demographic and encounter-level information. We use it to evaluate thoracic disease prediction and intersectional subgroup reliability based on available patient-level attributes. 

\textbf{Results.}
At 90\% nominal coverage, Table~\ref{tab:ham10000_additional}shows that \modelname achieves 90.30\% coverage and the lowest intersectional CovGap of 5.99 on HAM10000. Its average set size is 1.36, comparable to Marginal CP and Partial, and smaller than AFCP and FaReG.
On MIMIC-CXR (as shown in Table~\ref{tab:mimiccxr_additional}), \modelname achieves 90.25\% coverage with an average set size of 1.39, matching Marginal CP. Its intersectional CovGap of 1.21 is close to those of Marginal CP and AFCP, at 1.24 and 1.14, respectively, and substantially lower than FaReG at 3.78. AFCP achieves a slightly lower CovGap, but with larger prediction sets.

Together, \modelname achieves the lowest intersectional CovGap on HAM10000 and competitive subgroup reliability on MIMIC-CXR, while maintaining compact prediction sets on both datasets. 
These results extend our evaluation to dermatology and chest radiography, suggesting that the findings on the original benchmarks are not limited to a single clinical task or imaging modality.

\subsection{Sensitivity to Hyperparameters and Subgroup Configurations}
\label{subsec:exp_sensitivity}


\begin{figure}[t]
  \centering
  \begin{minipage}[t]{0.24\linewidth}
    \centering
    \includegraphics[width=\linewidth]{pdf_syn.pdf}
    {\small (a) K=3}
  \end{minipage}
  \hfill
  \begin{minipage}[t]{0.23\linewidth}
    \centering
    \includegraphics[width=\linewidth]{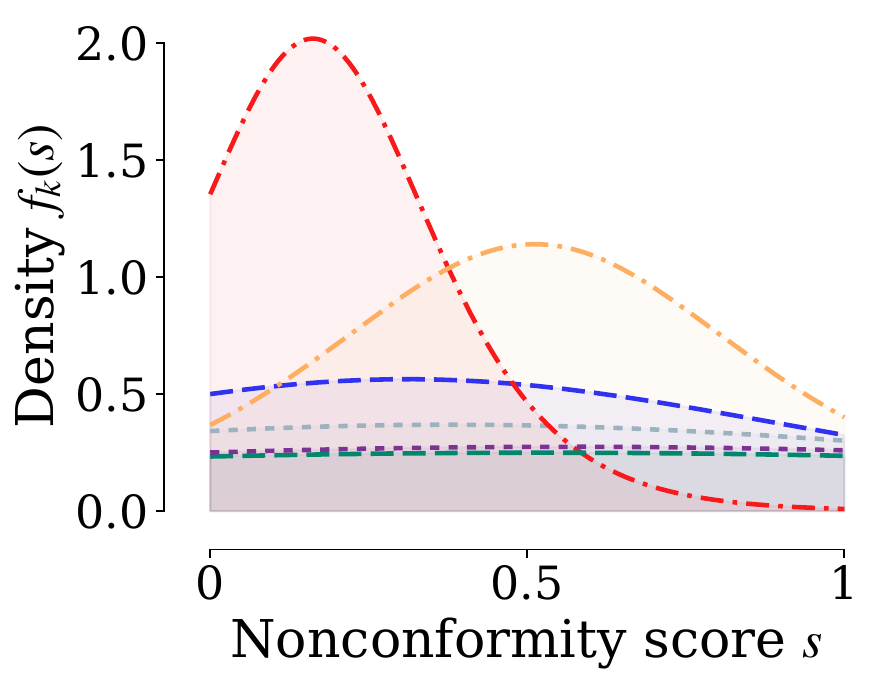}
    {\small (b) K=6}
  \end{minipage}
  \hfill
  \begin{minipage}[t]{0.24\linewidth}
    \centering
    \includegraphics[width=\linewidth]{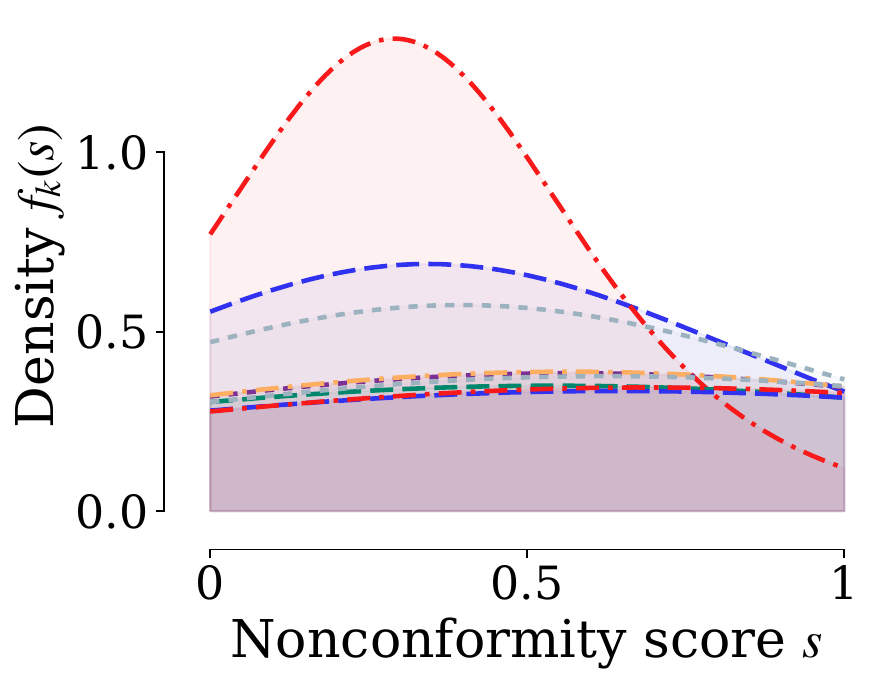}
    {\small (c) K=9}
  \end{minipage}
  \hfill
  \begin{minipage}[t]{0.24\linewidth}
    \centering
    \includegraphics[width=\linewidth]{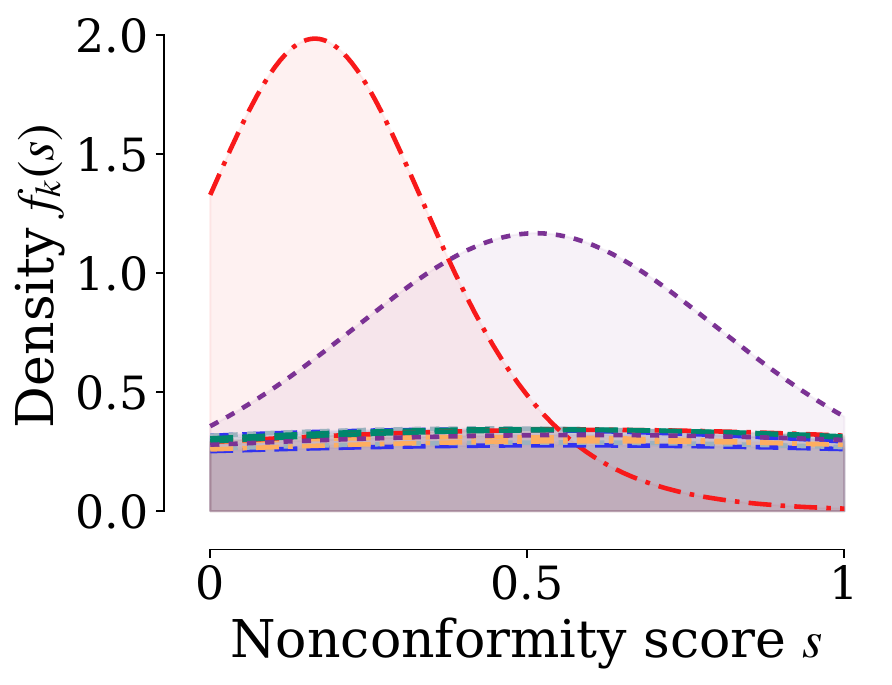}
    {\small (d) K=12}
  \end{minipage}
    \caption{
    Effect of $K$ on learned latent component densities for SYN. Increasing $K$ refines the score law components, but large $K$ introduces overlapping or redundant densities.}
    \vspace{-2mm}
  \label{app_pdfK_syn}
\end{figure}

\begin{figure}[t]
  \centering
  \begin{minipage}[t]{0.24\linewidth}
    \centering
    \includegraphics[width=\linewidth]{pdf_mimic.pdf}
    {\small (a) K=3}
  \end{minipage}
  \hfill
  \begin{minipage}[t]{0.23\linewidth}
    \centering
    \includegraphics[width=\linewidth]{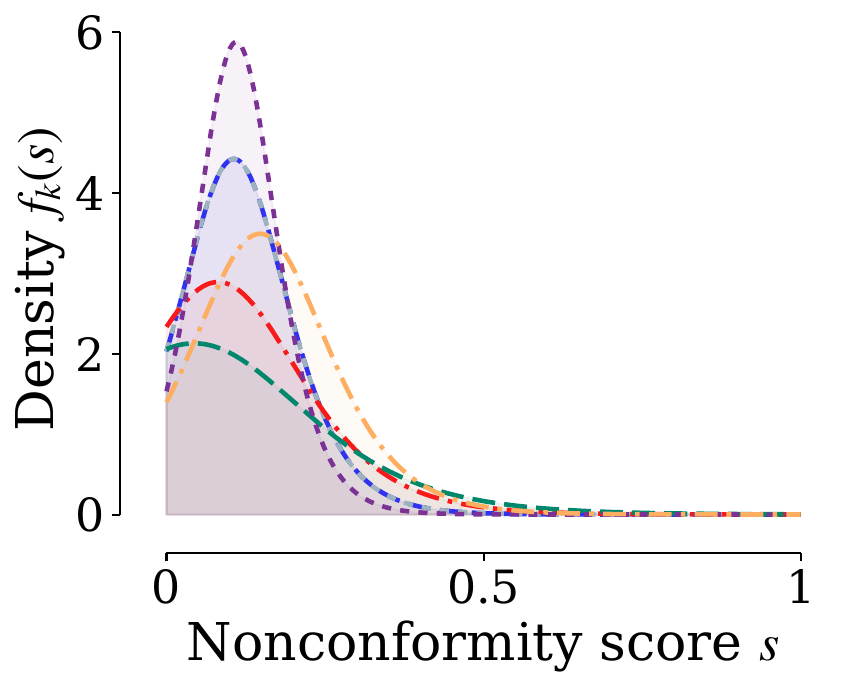}
    {\small (b) K=6}
  \end{minipage}
  \hfill
  \begin{minipage}[t]{0.24\linewidth}
    \centering
    \includegraphics[width=\linewidth]{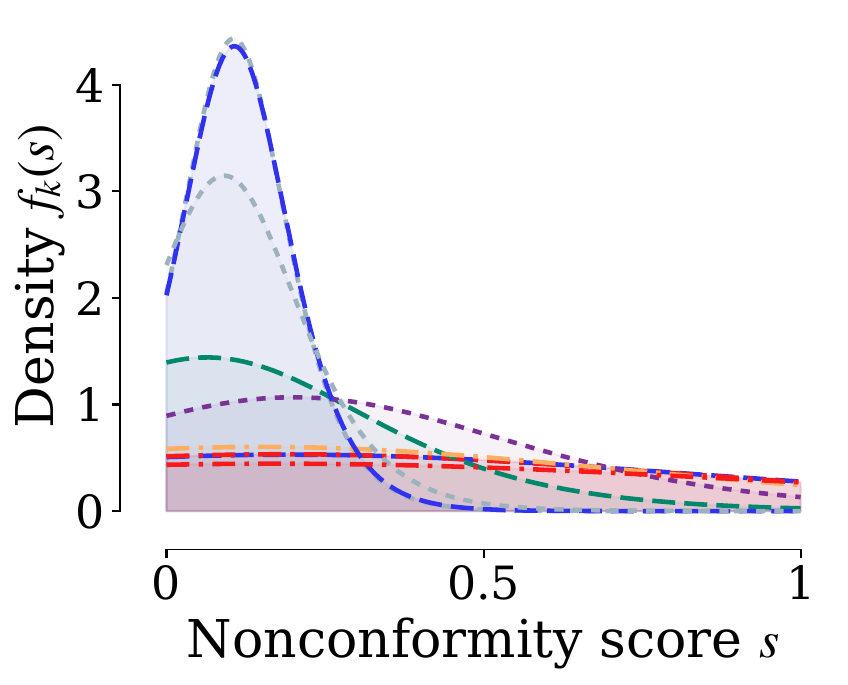}
    {\small (c) K=9}
  \end{minipage}
  \hfill
  \begin{minipage}[t]{0.24\linewidth}
    \centering
    \includegraphics[width=\linewidth]{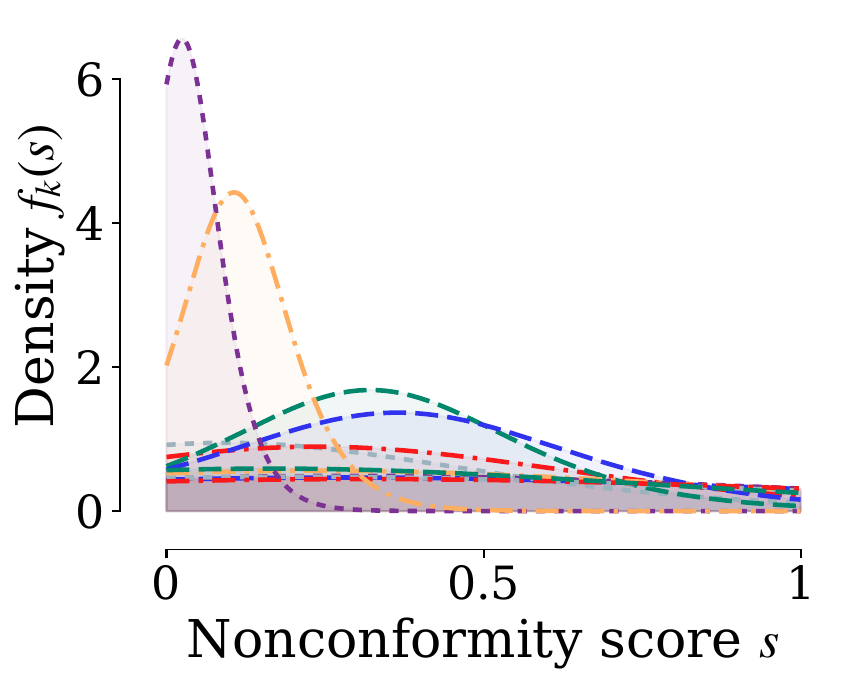}
    {\small (d) K=12}
  \end{minipage}
    \caption{
    Effect of $K$ on learned latent component densities for MIMIC-IV. Larger $K$ refines the dominant low score region and tail behavior, but also introduces increasingly overlapping components.}
    \vspace{-2mm}
  \label{app_pdfK_mimic}
\end{figure}

\subsubsection{Effect of sample sparsity}
Tables~\ref{tab_syn_sample}–\ref{tab_bach_sample} study how performance changes as the effective sample size varies.
As the effective sample size decreases, methods based on explicit subgroup equalization or localization tend to become increasingly conservative, while global calibration remains efficient but suffers from large subgroup disparity.
By contrast, \modelname degrades more gracefully in the sparse regime and benefits more consistently from additional data as sample size grows.
This pattern is strongest on the synthetic benchmark and on BACH, where subgroup heterogeneity is more structured, and remains visible on MIMIC-IV, where \modelname is particularly competitive in the sparse and moderate sample regimes.

For a clearer comparison, we also provide separate box plots for each metric across methods and sample sizes. 
Figures~\ref{fig_coverage_hist}, \ref{fig_size_hist}, and \ref{fig_covgap_hist} show the distributions of overall coverage, average prediction set size, and subgroup coverage gap under varying sample sparsity. 
Each box summarizes 20 independent trials with different random seeds for a given dataset and sample-size setting. 
Across datasets, \modelname remains near the nominal target with competitive set sizes and consistently low subgroup disparity, especially in sparse regimes where competing methods become more conservative or more variable.

\subsubsection{Effect of the number of latent components}
Figure~\ref{app_pdfK_syn} visualizes the learned component densities on SYN under different choices of $K$.
When $K$ is small, the model learns a few distinct score law patterns that separate low score and high score calibration regimes.
As $K$ increases, additional components tend to overlap with existing ones or become nearly flat, suggesting that the dominant calibration structure has already been captured by a smaller number of components.
This behavior indicates that $K$ controls the granularity of stochastic sharing rather than simply increasing model capacity.
In practice, overly large $K$ may introduce redundant components, while a moderate $K$ provides a compact set of reusable score law building blocks for constructing sample dependent local score laws.

Figure~\ref{app_pdfK_mimic} shows a similar sensitivity pattern on MIMIC-IV, but with a different score structure.
Most learned components concentrate near small nonconformity scores, reflecting the strong mass of the empirical score distribution in the low score region.
Increasing $K$ refines this region and introduces broader components that capture the remaining tail behavior.
However, when $K$ becomes large, several components overlap substantially or become weakly differentiated.
This suggests that additional components mainly refine existing calibration patterns rather than uncover entirely new score regimes.
Together, the SYN and MIMIC-IV results show that a moderate number of latent components is sufficient to capture the dominant score law structure, while larger $K$ may lead to redundant calibration sharing.

\begin{figure}
    \centering
    \includegraphics[width=1\linewidth]{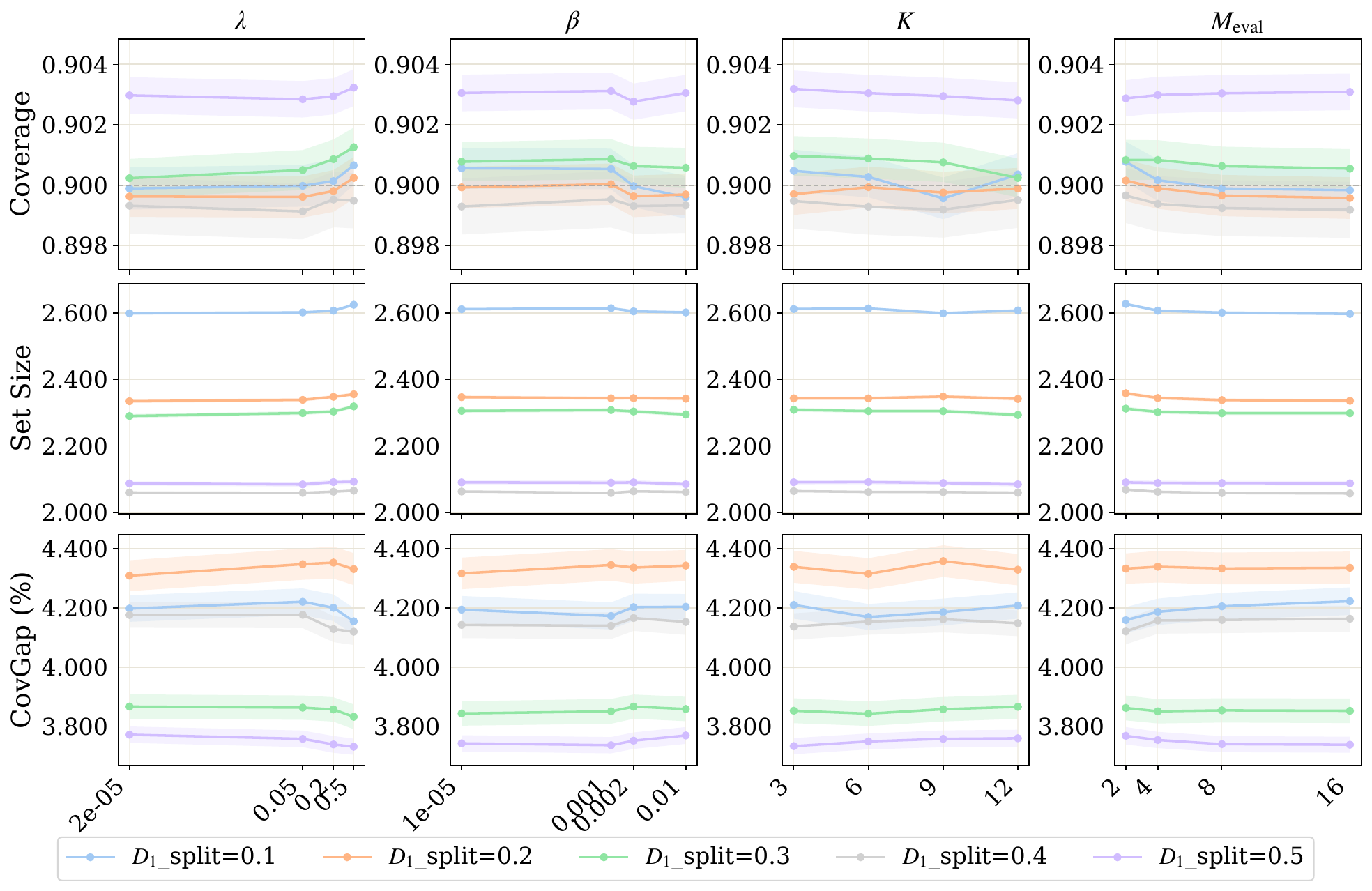}
    \caption{Hyperparameter sensitivity of \modelname on SYN. We vary $\lambda_{\mathrm{nll}}$, $\lambda_{\mathrm{bal}}$, $K$, and $M_{\mathrm{eval}}$, and report coverage, average set size, and subgroup coverage gap under different $D_1$ split ratios. Shaded bands show one standard deviation. \textbf{Takeaway:} \modelname is largely insensitive to these hyperparameters, while larger samples for score-law learning reduce covgap.}
    \label{fig_syn_sensitivity}
\end{figure}

\begin{figure}
    \centering
    \includegraphics[width=1\linewidth]{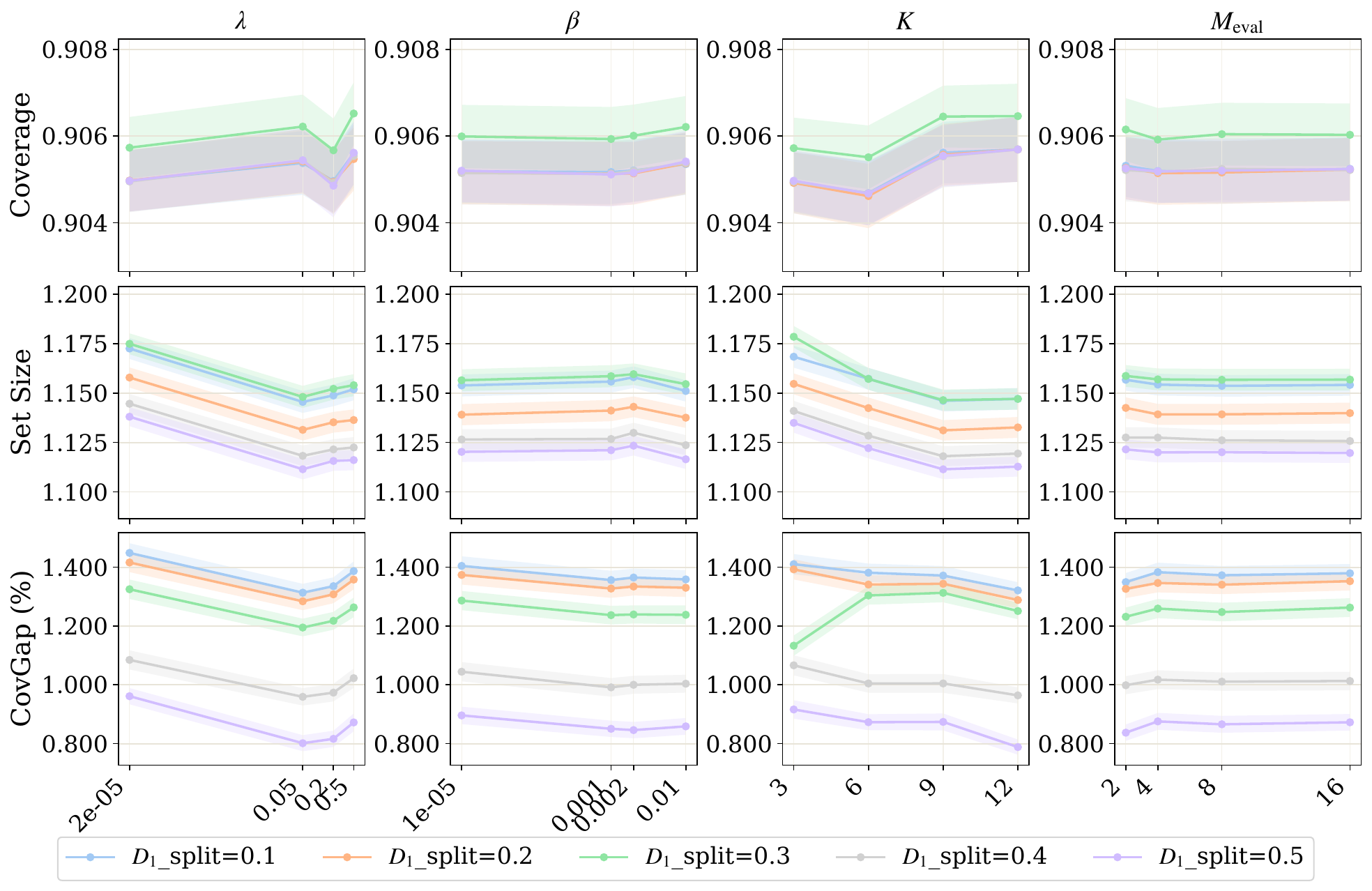}
    \caption{Hyperparameter sensitivity of \modelname on MIMIC-IV (ICU delay). \textbf{Takeaway:} \modelname is largely insensitive to these hyperparameters. Coverage remains close to the nominal level, and most of the variation in efficiency and subgroup reliability is driven by the $D_1$ split ratio rather.}
    \label{fig_mimic_sensitivity}
\end{figure}

\subsubsection{Hyperparameter sensitivity}
Figures~\ref{fig_syn_sensitivity} and \ref{fig_mimic_sensitivity} examine the sensitivity of \modelname w.r.t. several key hyperparameters, including $\lambda$, $\beta$, the number of latent calibration components $K$, and the number of evaluation latent samples $M_{\mathrm{eval}}$, under different $D_1$ split ratios. Across both datasets, the hyperparameter curves are generally flat, indicating that the performance of \modelname does not rely on delicate tuning. 

On the SYN, this stability is particularly clear: coverage remains close to the nominal target across all tested settings, while the changes in average prediction set size and subgroup coverage gap are modest. The most visible effect comes from the $D_1$ split ratio, where allocating more data to score law learning consistently improves subgroup reliability and reduces prediction set size. 
On MIMIC-IV, it show a similar pattern. Coverage remains tightly concentrated around the nominal level across all tested settings, with modest changes in AvgSize and CovGap. Compared with hyperparameter choices, allocation training data to local score law learning has a more visible effect on the final tradeoff.
Overall, these results suggest that the gains of \modelname stem primarily from its structured local score law design rather than from aggressive hyperparameter tuning.

\begin{figure}
    \centering
    \includegraphics[width=0.8\linewidth]{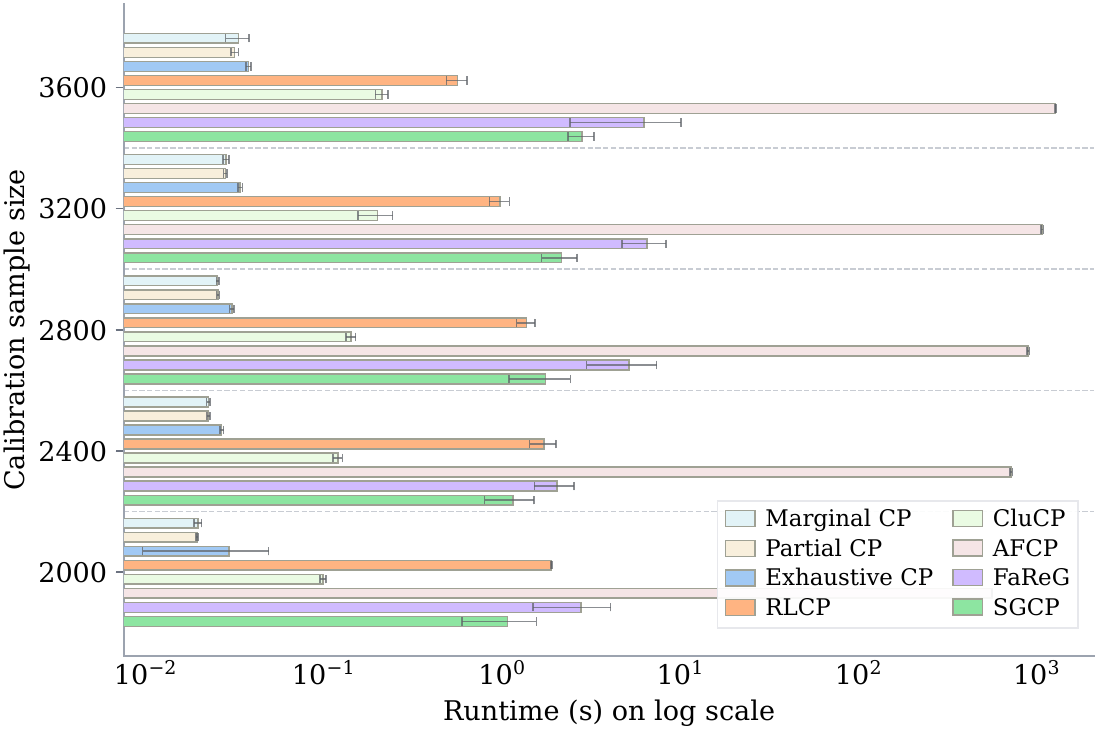}
    \caption{Wall clock time comparison w.r. calibration sample. We report the wall-clock time for each method, averaged over repeated runs; error bars indicate one standard deviation. The horizontal axis is shown on a log scale. \textbf{Takeaway:} \modelname has moderate computational cost and is faster than AFCP, while remaining competitive with other subgroup aware baselines across all sample sizes.}
    \vspace{-2mm}
    \label{appendix_runtime}
\end{figure}

\subsection{Wall-Clock Time}\label{app_walltime}
We compare the wall-clock time of the full \modelname pipeline with all baselines under varying calibration sample sizes. 
Figure~\ref{appendix_runtime} reports the average runtime over repeated runs, with error bars indicating one standard deviation. 
Marginal CP and Partial CP are the fastest methods, as they require only global calibration. 
RLCP and \modelname incur additional cost from sample-dependent calibration, but remain moderate and scale smoothly with sample size. 
AFCP is the most expensive baseline by a large margin, reflecting its repeated adaptive subgroup selection and conformal recalibration, while CluCP and FaReG lie in between. 
Overall, \modelname provides a favorable computational tradeoff, remaining practical across all tested sample sizes while achieving stronger subgroup reliability than simpler global or rigid group based baselines.

\subsection{Limitations}\label{app_limitation}
Our study has several limitations. All experiments are conducted on retrospective benchmarks, so prospective validation is needed to understand the practical impact of subgroup reliable prediction sets in real clinical workflows. In addition, our current framework is developed for multiclass classification and does not address regression or other prediction settings. Extending the method beyond classification and evaluating it under deployment level constraints are left to future work.

\section{Supplement to Related Work}\label{appendix_rw}

\paragraph{Group-conditional and fairness-aware conformal prediction.}
A substantial body of work studies fairness-aware conformal prediction through coverage parity across groups that are either specified in advance or adaptively selected from data. 
Vadlamani et.al~\cite{DBLP:conf/iclr/VadlamaniSMP025} develop a unified post-hoc framework that extends classical fairness notions to prediction sets by adjusting conformity thresholds in a fairness-aware manner. 
AFCP~\cite{10.5555/3737916.3741370} study conformal classification with equalized coverage for adaptively selected groups, selecting features that reflect potential model limitations or biases in order to balance predictive efficiency and group reliability. 
FAREG~\citep{xu2026fair} goes beyond predefined sensitive groups by learning representation-based subgroups and adaptively promoting equalized coverage for populations that would otherwise receive unfair treatment. 
SAGCCI~\citep{10.5555/3780338.3781045} examines the tradeoff between fairness and efficiency by clustering protected groups with similar conformal score distributions and by using surrogate outcomes to improve threshold estimation. 
This is particularly close in spirit to our work in its use of score-distribution similarity to share calibration information. 
The key distinction lies in the granularity of the sharing mechanism. 
SAGCCI shares calibration information across clusters of observed protected groups, whereas \modelname learns a sample-dependent stochastic mixture over latent calibration components and thereby induces a local score law for each individual sample. 
Consequently, the calibration-sharing structure learned by \modelname need not align with protected-group labels and does not rely on surrogate outcomes.

\paragraph{Structured sharing and learned calibration groups.}
A related line of work improves upon a single global calibration rule by sharing calibration evidence across structured partitions. 
Clustered conformal prediction~\citep{10.5555/3666122.3668939} groups labels with similar score distributions in many-class settings and performs calibration at the cluster level, thereby interpolating between standard conformal prediction and fully class-conditional conformal prediction. 
Conceptually similar ideas also arise beyond conformal prediction. 
For instance, VIR~\citep{DBLP:conf/nips/WangW23} smooths probabilistic uncertainty estimates across neighboring label regions in imbalanced regression. 
Together, these works underscore the value of structured sharing: calibration and uncertainty estimation can often benefit from borrowing information across statistically related regions, rather than treating all samples globally or calibrating each group in isolation.
Another line of work enlarges or learns the group structure used for conditional calibration. 
Kandinsky Conformal Prediction~\citep{bairaktari2025kandinsky} generalizes group-conditional conformal inference to overlapping and fractional group functions defined on $(X,Y)$, thereby expanding the collection of groups over which weighted coverage deviations can be controlled. 
Recent work on identifying homogeneous and interpretable groups for conformal prediction likewise uses nonconformity-score behavior to learn explicit partitions of the input space before applying group-conditional calibration~\citep{martinez2024identifying}. 
These methods are closely related to ours in their shared emphasis on calibration-relevant structure beyond predefined sensitive attributes.

Our work adopts a complementary perspective. 
Instead of specifying a family of group functions or assigning each sample to a hard discovered partition, \modelname learns a sample-dependent local score law through stochastic mixing over shared latent calibration components. 
This enables calibration information to be shared according to learned score behavior, while the final prediction set is still obtained through standard split conformal calibration applied to transformed scores.

\end{document}